\documentclass[conference]{IEEEtran}

\usepackage{amsmath,amssymb,amsfonts}
\usepackage{graphicx}
\usepackage{booktabs}
\usepackage{multirow}
\usepackage{tabularx}
\usepackage{makecell}
\usepackage{xcolor}
\usepackage{fontawesome5}
\usepackage{cuted}
\usepackage{caption}
\usepackage{tikz}
\usepackage{caption}
\usetikzlibrary{arrows.meta,positioning,fit,backgrounds,calc}
\usepackage[numbers,sort&compress]{natbib}
\usepackage[colorlinks=true,linkcolor=blue!60!black,citecolor=blue!60!black,urlcolor=blue!60!black]{hyperref}
\usepackage[capitalize]{cleveref}
\usepackage{microtype}
\usepackage{float}

\newcommand{\model}{\mbox{FACT}}
\newcommand{\tbd}{\textcolor{gray}{--}}
\newcommand{\stageone}{Track-Guided Coarse Planner}
\newcommand{\stagetwo}{Force--Action Joint Refiner}

\newcommand{\best}[1]{\textbf{#1}}
\newcommand{\algabbr}{\model{}}

\title{\model{}: A Multimodal Force--Action--Track\\ Foundation Model for Human-to-Robot Dexterity}
\author{%
\IEEEauthorblockN{Zhi Cao$^{1,2*}$, Yining Hong$^{2*}$, Jiaolong Yang$^{3}$, Yejin Choi$^{2}$, Li Fei-Fei$^{2}$, Jiajun Wu$^{2\dagger}$, Huang Huang$^{2\dagger}$}
\IEEEauthorblockA{$^{1}$University of Michigan \quad $^{2}$Stanford University \quad $^{3}$Microsoft Research\\
$^{*}$Equal contribution \quad $^{\dagger}$Equal advising}\\[0.6em]
\faGlobe\ \textbf{Website:}  \url{https://fact-1-0.github.io/}
}

\begin{document}
\maketitle

\begin{strip}
\centering
\begin{minipage}{\textwidth}
    \centering
    \vspace{-2em}
    \includegraphics[width=\textwidth]{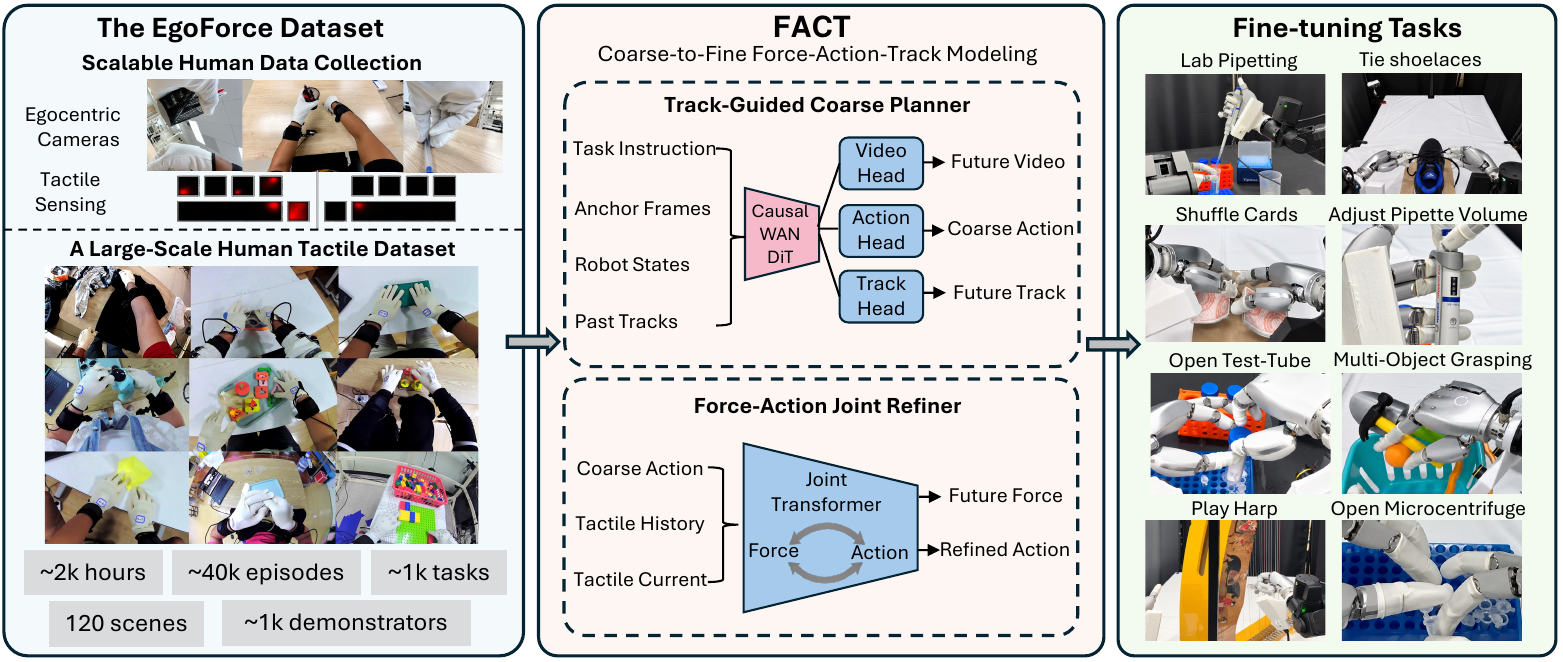}
    \captionof{figure}{
\model{} combines large-scale force-rich human experience with robot demonstrations for pretraining. Its coarse-to-fine architecture uses visual
tracks and video prediction to shape coarse action, followed by joint
force--action refinement for fine-grained execution. The pretrained model
adapts to diverse tasks, from tool use,
musical instruments, to precision laboratory procedures.\looseness=-1
    }
    \label{fig:teaser}
\end{minipage}
\end{strip}

\begin{abstract}

Human demonstrations offer a scalable alternative to costly robot data collection for dexterous manipulation, but most human datasets lack force measurements. Dexterous manipulation requires not only visual and proprioceptive understanding but also the ability to anticipate and regulate interaction forces. To scale force-aware human--robot co-training, we introduce \emph{EgoForce}, a large-scale human tactile dataset capturing force-rich manipulation experience, and combine it with robot demonstrations for pretraining. We present \algabbr{}, a robot foundation model that cascades from prospective visual planning to force-aware refinement for dexterous manipulation. \algabbr{} comprises two jointly trained components. A Track-Guided Coarse Planner jointly predicts actions, future video, and visual tracks. Future video and visual tracks also provide a shared task representation that facilitates cross-embodiment knowledge transfer and coarse action generation. Building on this coarse plan, a Force--Action Joint Refiner jointly predicts future force and refines the coarse action, allowing anticipated force and action to shape each other during refinement. We evaluate the model on ten tasks requiring fine-grained coordination and force feedback, including tool use, musical instruments, and precision laboratory procedures.  Using as few as 10 robot trajectories with 90 human trajectories for fine-tuning, \algabbr{} outperforms both baselines, with success-rate gains of up to 60 percentage points over T-Rex. Ablations show that force supervision in human pretraining data is crucial for downstream robot performance after fine-tuning, while visual track prediction and training with future-force prediction improve dexterous execution. Experiments varying the human--robot data ratio during fine-tuning further suggest that \model{} learns from human demonstrations nearly as effectively as from robot demonstrations.\looseness=-1
\end{abstract}

\section{Introduction}
\label{sec:intro}


Collecting diverse robot demonstrations for dexterous manipulation at scale is expensive, making human demonstrations an attractive source of manipulation experience. Most human manipulation datasets capture vision and motion without measuring contact forces. However, contact-rich dexterous manipulation often requires both prospective visual planning and force-aware refinement~\cite{todorov2002optimal,scott2004optimal}: where the hands go is guided by vision and task context, while how the fingers finish the motion is shaped by contact forces. The former determines the coarse structure of the motion; the latter governs its fine-grained execution. Force measurements can provide supervision for how hand motion and physical interaction evolve together, allowing human experience to better support force-aware robot learning.\looseness=-1

To support force-aware learning at scale, we develop a collection platform combining three egocentric views, bimanual tactile gloves, and automated force calibration to capture synchronized vision, hand motion, and tactile force during natural manipulation. Using this platform, we collect EgoForce, a large-scale human tactile dataset comprising approximately 2k hours of manipulation across 40k episodes, 1k tasks, and 1k demonstrators.\looseness=-1

To transfer this force-rich human experience to robots, we introduce \model{}, a dexterous foundation model that
cascades from prospective visual motion planning to force-aware refinement for human-to-robot dexterity (\Cref{fig:teaser}). The model comprises two jointly trained components for coarse motion planning and fine-grained force-aware execution.
The Track-Guided Coarse Planner jointly predicts future video, visual tracks, and an action chunk. Building on the coarse plan, a Force--Action Joint Refiner takes the partially denoised action together with observed tactile signals and co-denoises the remaining action and future fingertip force through bidirectional self-attention, allowing action and force to reciprocally condition one another for action refinement. Future video and visual track prediction not only ground coarse action generation in anticipated visual motion but also provide an embodiment-agnostic task representation that encourages knowledge transfer between humans and robots.\looseness=-1

We pretrain \model{} on our large-scale EgoForce dataset together with a small set of robot
demonstrations. We then fine-tune and evaluate it across 10 real-world
tasks, spanning everyday skills such as tying shoelaces, precision laboratory
procedures such as pipetting and microcentrifuge-tube manipulation, as well as
tool use and musical instruments. Our ablations show that force supervision in human pretraining data provides the largest gains on tasks demanding precise contact and force control, while visual track prediction and training with future-force prediction improve dexterous execution. Experiments with different human--robot data ratios during fine-tuning suggest that human demonstrations can contribute nearly as much as robot demonstrations to policy learning, supporting human experience as a scalable source of dexterous manipulation skills.\looseness=-1


Our contributions are threefold:
\begin{itemize}
    \item We introduce \model{}, a dexterous foundation model that cascades from track-based visual motion planning to force-aware refinement for dexterous manipulation.

    \item We introduce EgoForce, a large-scale human tactile dexterous manipulation dataset and scalable collection platform for pretraining.

\item Across ten real-world tasks, we show the benefits of force supervision in human pretraining data and visual track and future-force prediction during training, together with the effectiveness of \model{} in using human data for fine-tuning.\looseness=-1
\end{itemize}


\section{Related Work}
\label{sec:related}
\begin{figure*}[htbp]
\centering
\includegraphics[width=\textwidth]{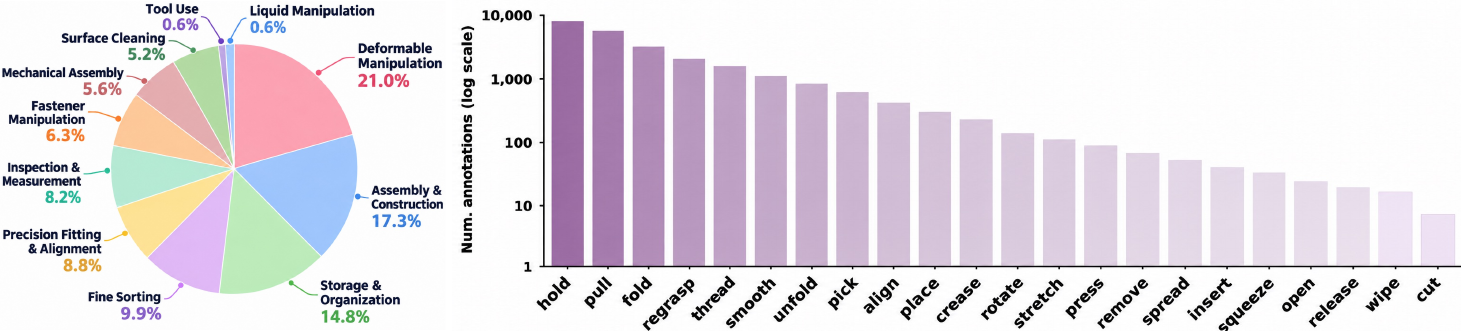}
\caption{ \textbf{\emph{EgoForce} task and skill distribution.}
Left: episode share by task category, with over two thirds of the corpus
covering contact-rich fine manipulation across diverse task categories.
Right: atomic-skill annotations on a log scale, showing a broad distribution
of dexterous, force-dependent skills beyond simple transport.}
\vspace{-1.5em}
\label{fig:datastats}
\end{figure*}

Large egocentric corpora
\cite{grauman2022ego4d,damen2018epic} have enabled robot learning from human
experience through pretrained visual representations~\cite{nair2022r3m},
imitation from human play~\cite{wang2023mimicplay}, portable in-the-wild
capture systems~\cite{chi2024umi,wang2024dexcap}, egocentric co-training
\cite{kareer2024egomimic,hoque2025egodex}, and human--robot mixtures in VLA
pretraining~\cite{trex}. These sources are largely force-blind, capturing hand motion without contact forces.
\emph{EgoForce} (\Cref{sec:egoforce}) closes this gap with 2{,}000 hours of human manipulation paired with calibrated tactile force.

Learning from this experience requires policies that use touch to guide dexterous execution. Tactile sensing has supported robot manipulation through vision-based skins
\cite{yuan2017gelsight,wu2024threedvitac}, learned touch representations
\cite{higuera2024sparsh,guzey2023tdex}, in-hand control~\cite{qi2023general},
and tactile-aware imitation and VLA policies
\cite{heng2025vitacformer,huang2025tactilevla}. Related work has explored
force-centric demonstration capture~\cite{hu2024forcemimic} and auxiliary
future contact or force prediction~\cite{li2024forceaware}. 
Hierarchical manipulation further separates planning from fine-grained control through
explicit action hierarchies~\cite{belkhale2023hydra,belkhale2024rth}, staged
VLA post-training~\cite{black2025pi06}, and slow-fast tactile refinement
\cite{xue2025rdp,trex}. T-Rex~\cite{trex} combines low-frequency visuomotor action generation with asynchronous tactile refinement, conditioning coarse actions on vector-quantized force history~\cite{vandenoord2017vqvae}. Force remains an input, leaving action--contact dynamics implicit and refinement dependent on the contact information preserved by compression. \model{} instead jointly denoises continuous future force and action, allowing them to shape one another during refinement. 

Future video and visual track prediction support human-to-robot transfer through a shared representation of anticipated scene motion across different action spaces. Work coupling action generation with future visual prediction includes video pretraining for
manipulation~\cite{wu2023gr1}, policy-as-video~\cite{du2023unipi}, and
interactive world simulators~\cite{yang2024unisim,agarwal2025cosmos}.
Visual point tracks provide an explicit, embodiment-agnostic description of scene motion. They have served as policy
interfaces~\cite{wen2023atm}, transfer media from human video
\cite{bharadhwaj2024track2act,xu2024flow,cao2026tractbridgingrobotcontrol}, and cross-domain motion
representations~\cite{xu2024flow}, enabled by general-purpose trackers
\cite{karaev2024cotracker}. Similar to~\cite{cao2026tractbridgingrobotcontrol}, \model{} jointly predicts actions, future video, and visual tracks in its Wan2.2-based coarse planner~\cite{wan2025}. These shared prediction targets let human demonstrations supervise motion representations that ground robot action generation in anticipated visual motion.\looseness=-1

\section{The EgoForce Dataset}
\label{sec:egoforce}

Human demonstrations provide a scalable source of dexterous manipulation
experience, but existing human data captures vision without
the forces involved in interaction, while collecting force-supervised
data at comparable scale on robots is expensive and difficult. We therefore
develop \emph{EgoForce}, a large-scale human tactile dataset paired with a
collection platform that captures synchronized vision, hand motion,
and calibrated tactile force for \model{} pretraining.

\paragraph{Scalable multimodal collection platform}
We design a wearable platform to capture natural bimanual manipulation while
minimally constraining the demonstrator. Each participant wears a pair of
tactile gloves together with three egocentric RGB cameras mounted on the head
and both wrists. The platform synchronizes the visual streams with glove motion
and tactile force measurements, producing temporally aligned multimodal
trajectories across diverse tasks and environments.

\paragraph{Automatic force calibration}

To obtain comparable force measurements across sensing elements, gloves, and collection sessions, we develop an automatic calibration suite that maps raw A/D readings to physical force. A pneumatic bladder applies approximately uniform pressure across the tactile surface, swept from $0$ to $50$ psi, while a reference pressure sensor records pressure synchronously with glove readings. For each sensing element, we compute force as $F_i=P_iA_i$, where $P_i$ is the reference pressure and $A_i$ the sensing area. These paired A/D and force measurements yield a per-element calibration mapping applied during data collection.

\paragraph{The EgoForce dataset}
Using this platform, we collect \emph{EgoForce}, comprising approximately
2,000 hours of human demonstrations across 40k episodes, 1,000 tasks, 120
scenes, and 1,000 demonstrators. Each demonstration contains synchronized
multi-view egocentric RGB observations, human hand motion, and calibrated
tactile force signals, capturing both manipulation behavior and the physical
contact underlying it. 

\paragraph{Task distribution: dexterous by construction}
\Cref{fig:datastats} shows that \emph{EgoForce} is deliberately skewed toward
dexterous, contact-rich interaction rather than pick-and-place transport.
The largest categories are deformable manipulation (21.0\%), assembly
\& construction (17.3\%), and storage \& organization (14.8\%); together
with fine sorting, precision fitting, fastener manipulation, and mechanical
assembly, fine manipulation comprises over two thirds of the data. At the
verb level, after \emph{hold} and \emph{pull}, frequent annotations include
\emph{fold}, \emph{regrasp}, \emph{thread}, \emph{smooth}, and
\emph{crease}, reflecting in-hand reorientation and deformable skills
whose success depends on contact forces, while \emph{pick} and \emph{place}
rank only eighth and tenth. Similar visual states, such as a held fold or
threaded lace, can require very different force profiles, motivating calibrated force supervision. 

\begin{figure*}[t]
\centering
\includegraphics[width=0.95\textwidth]{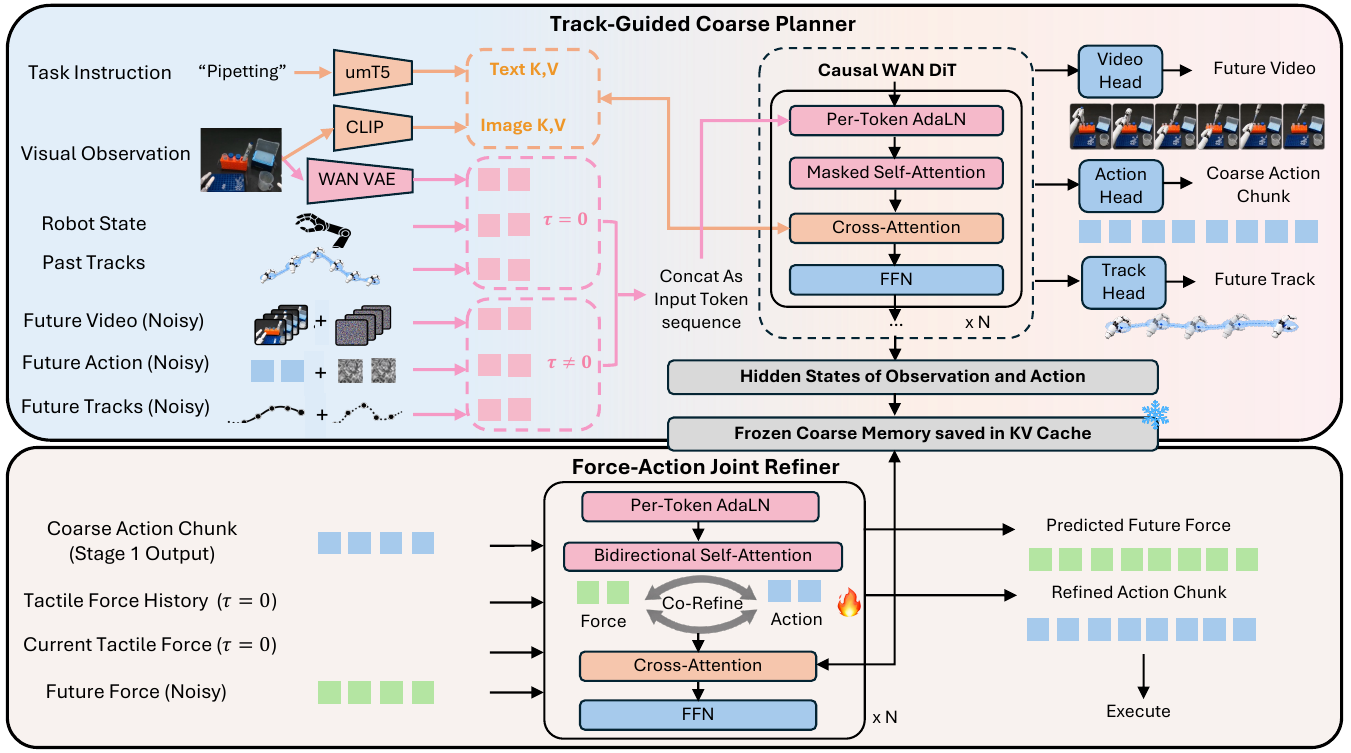}
\caption{\textbf{\model{} architecture.} The \stageone{} jointly denoises
future video, visual tracks, and a coarse action chunk with a block-causal
video DiT; per-token noise conditioning lets clean context and noisy targets
coexist. At the $\tau{=}0.4$ handoff, video and tracks are read out and the
coarse hidden states are frozen. The \stagetwo{} reads them through cached
cross-attention and co-denoises the remaining action ($\tau_a\!:0.4{\to}0$)
with continuous future fingertip force ($\tau_f\!:1{\to}0$), using observed
tactile signals as clean context.}
\vspace{-1em}
\label{fig:arch}
\end{figure*}

\vspace{-0.5em}
\section{The \model{} Framework}
\label{sec:method}

\model{} is a flow-matching model for transferring dexterous manipulation
from large-scale human experience to robot execution. \model{} cascades from
prospective visual planning to force-aware refinement (\Cref{fig:arch}) through
two jointly trained components.
The \stageone{} jointly
predicts future video, visual tracks, and actions. Future video and tracks
provide a shared representation of task progress across human and robot
embodiments, allowing human experience to shape coarse robot motion. The \stagetwo{} jointly refines the action suffix and predicts future fingertip
force while cross-attending to the coarse representation from the \stageone{}. 
This
separation supports cross-embodiment learning through shared visual motion and
fine-grained robot execution through force--action co-prediction. 
We pretrain \model{} on our \emph{EgoForce} dataset together with 100 hours of robot
teleoperation demonstrations from T-Rex~\cite{trex}.

At inference time, the policy receives the current visual observation,
robot state, task instruction, visual track history, and tactile history.
Given these observations, \model{} outputs a 16-step, 20\,Hz action chunk $a\in\mathbb{R}^{16\times62}$, consisting of delta-base pose and absolute
hand joints for each arm, and the prediction of future force at the chunk's 5\,Hz refresh slots,
$\hat F\in\mathbb{R}^{4\times10\times6}$, as well as the future visual track and video prediction.\looseness=-1

\subsection{\stageone{}}
\label{sec:stage1}

The \stageone{} learns a coarse visuomotor plan by jointly predicting future video, visual tracks, and actions, which is instantiated with a block-causal video DiT that autoregressively models the interaction over time.\looseness=-1


\paragraph{Multimodal representation}
As illustrated in \Cref{fig:arch}, the \stageone{} separates its inputs into three
groups. Task instruction and the anchor frame, encoded by umT5-XXL~\cite{chung2023unimaxfairereffectivelanguage} and CLIP~\cite{radford2021learningtransferablevisualmodels},
provide cross-attention conditioning. Robot state, past visual tracks, and
conditioning video latents provide clean context, while future video latents,
actions, and future visual tracks are jointly denoised.
The model operates autoregressively over temporal blocks containing video,
action, state, and past and future visual-track tokens. Additional clean
observation tokens condition the autoregressive sequence, while text and CLIP
features remain outside the self-attention sequence and enter only through
cross-attention.\looseness=-1

\paragraph{Per-token flow matching}
The \stageone{} jointly generates future video, actions, and tracks through flow
matching. For each generated modality with clean target $c$ and Gaussian noise
$\varepsilon$, we construct
\begin{equation}
    x_\tau=(1-\tau)c+\tau\varepsilon,
    \qquad
    v^*=\varepsilon-c,
    \label{eq:stage1_flow}
\end{equation}
and train the model to predict the corresponding velocity field.

Each token is conditioned on its own noise level $\tau$, with $\tau{=}0$ denoting clean data and $\tau{=}1$ pure noise. Sinusoidal embeddings of $\tau$ are projected to six AdaLN modulation vectors per DiT block. Conditioning latents, state, and past tracks remain at $\tau{=}0$, while future video latents, actions, and tracks use block-wise sampled noise levels $\tau(k)\!\sim\!\mathrm{Beta}(1.5,1)$. This per-token conditioning lets clean context and noisy targets coexist in one forward pass, enabling \stageone{} to stop action denoising at an intermediate noise level and \stagetwo{} to continue refinement.

\paragraph{Attention topology}
Each DiT block applies per-token AdaLN, masked self-attention, cross-attention, and a feed-forward network. Within a four-block causal window, observation and action tokens attend to visual observation, action, and track tokens in the current and preceding blocks, plus the current state token. State tokens remain local to their block; track tokens attend only to tracks, preserving a dedicated geometric pathway. Cross-attention incorporates text and CLIP features, and modality-specific heads decode future video, the coarse action chunk, and future tracks.

\paragraph{Training objective}
The \stageone{} jointly optimizes visual dynamics, action, and track
prediction:
\begin{equation}
\mathcal{L}_{\mathrm{coarse}}
=
\mathcal{L}_{\mathrm{dyn}}
+
\mathcal{L}_{\mathrm{act}}
+
\lambda_t
\left(
    \mathcal{L}^{\mathrm{flow}}_{\mathrm{trk}}
    + 5\,\mathcal{L}^{\mathrm{pos}}_{\mathrm{trk}}
    + 2\,\mathcal{L}^{\mathrm{smooth}}_{\mathrm{trk}}
\right),
\label{eq:stage1_loss}
\end{equation}
where $\lambda_t{=}20$. Track losses are visibility-masked with a
motion-weighted boost to emphasize informative trajectories. In addition to
the track flow objective, the position and smoothness terms are computed from
the one-step reconstruction
\begin{equation}
    \hat{x}_0=x_\tau-\tau\hat{v}.
\end{equation}

\paragraph{Coarse handoff to the \stagetwo{}}
At inference, the coarse planner jointly denoises future video, tracks, and action for
six Euler steps, stopping at the intermediate noise level $\tau{=}0.4$.
Future video and tracks are then decoded in closed form, while the action
remains partially denoised at $\tau_a{=}0.4$ and is passed to the \stagetwo{} for
further force-aware refinement.
The frozen hidden states of visual observation and action tokens form the coarse
representation passed to the refiner, where they are projected once and stored
as a cross-attention key--value cache.

\subsection{\stagetwo}
\label{sec:stage2}

The \stagetwo{} jointly refines the remaining action and predicts future fingertip force. At the coarse handoff, future video and
tracks are read out, while the action remains partially denoised at
$\tau_a{=}0.4$. Observed tactile signals provide the current contact context,
while future force is explicitly predicted alongside the action, allowing
force and action to reciprocally shape one another throughout refinement.\looseness=-1

\begin{figure}[t]
\centering
\includegraphics[width=\linewidth]{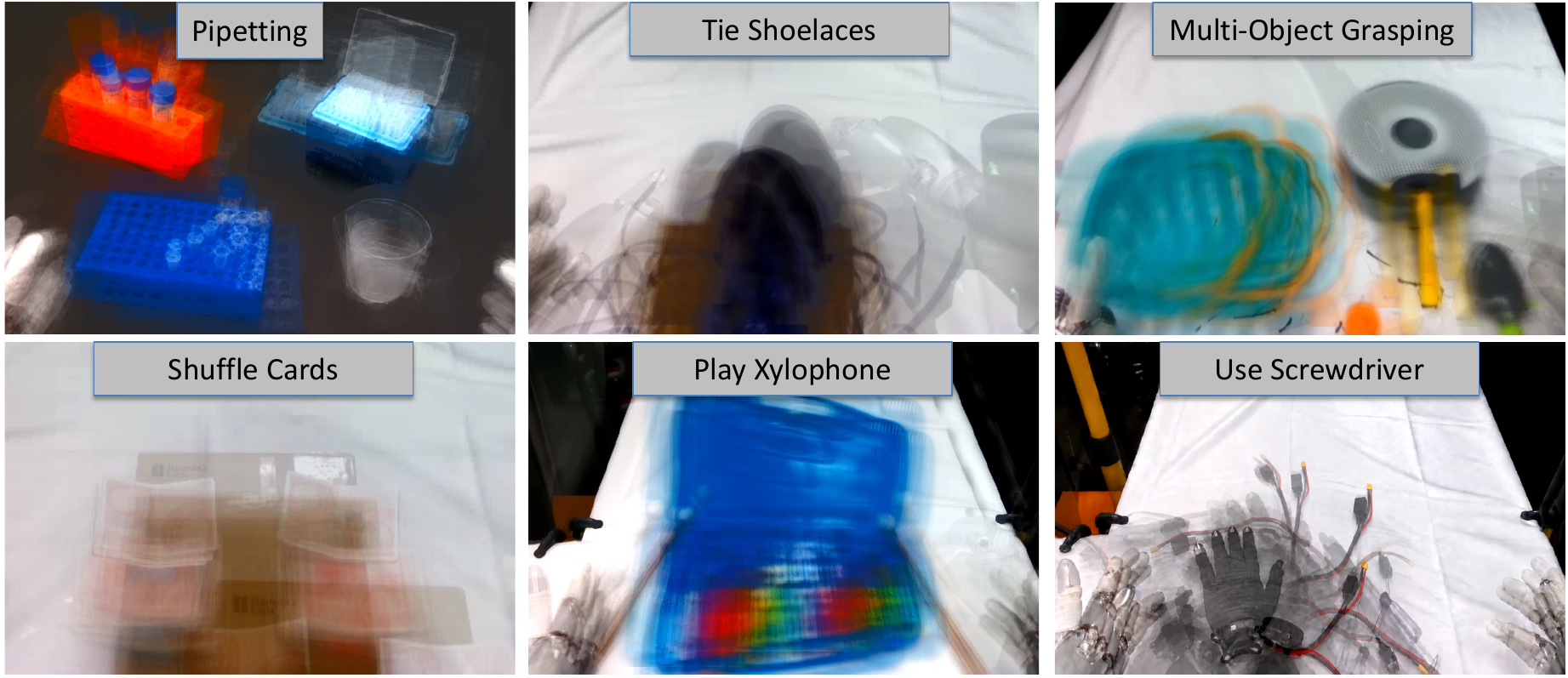}
\caption{\textbf{Evaluation randomization.} Overlaid initial frames from six evaluation tasks illustrate
rollout-level variation in object positions and poses.}
\label{fig:randomness}
\vspace{-2em}
\end{figure}

\paragraph{Force--action joint refinement}
The defining choice of the \stagetwo{} is that continuous future force is
a \emph{co-denoised prediction target, not an observation}: observed
fingertip force enters as clean context, while the future per-fingertip
force trajectory starts from pure noise at $\tau_f{=}1$ and is denoised
jointly with the action suffix---unlike T-Rex~\cite{trex}, where force
reaches the refiner only as vector-quantized observation codes and is
never predicted. Each refiner block applies per-token AdaLN($\tau$),
bidirectional self-attention, cross-attention to the coarse memory, and a
feed-forward network; bidirectional attention between the action suffix
and the noisy future-force tokens lets predicted force shape the evolving
action and the action shape its predicted force consequences at every
sampling step. Current tactile measurements thus describe the ongoing
physical interaction, while predicted future force represents the physical
consequence of the action being refined.

\paragraph{Coarse memory through cached cross-attention}
The frozen coarse hidden states provide the coarse visual plan through
cross-attention. At the $\tau{=}0.4$ handoff, the hidden states of observation
and action tokens are detached and retained, while track, state, and conditioning tokens are
dropped. These hidden states are projected once and stored as cross-attention
key--values. Because the coarse representation remains fixed after the handoff,
the cached key--values are reused across all four Euler steps and four tactile
refreshes. This allows repeated force--action refinement against the same coarse
visual plan without recomputing the \stageone{}.

\paragraph{Cascaded sampling}
Sampling performs four joint Euler steps with asymmetric schedules
$\tau_a\!:0.4{\to}0$ and $\tau_f\!:1{\to}0$. At each step, only the remaining
action suffix $a_{[o:16)}$ and unobserved future-force slots are updated; the
executed action prefix and observed force slots remain fixed. At the next
tactile refresh, the newly observed interaction is incorporated and the
remaining action suffix is refined again.

\paragraph{Training objective}
The \stagetwo{} jointly optimizes action refinement and future-force prediction:
\begin{equation}
\mathcal{L}_{\mathrm{refine}}=
\underbrace{\big\|\hat v_a-v_a^*\big\|^2_{[o:16)}}_{\text{action suffix}}
+\lambda_f
\underbrace{\big\|\hat v_F-v_F^*\big\|^2_{s>o/4}}_{\text{future force}}
+\mathcal{L}_{\mathrm{VQ}}.
\end{equation}
We sample
$\tau_a\!\sim\!\mathrm{Beta}(1.5,1)\!\cdot\!0.4$ and
$\tau_f\!\sim\!\mathrm{Beta}(1.5,1)$. The coarse memory forward is
evaluated at exactly $\tau{=}0.4$ using the same suffix noise draw, matching
the cascaded inference state during training. We use $\lambda_f{=}0.5$ in all
experiments.

\subsection{Implementation Details}
The \stageone{} is a 5B video DiT based on Wan2.2-TI2V (30 blocks, $d=3072$, 24 heads). Each $0.8$\,s block spans two latent video frames and contains 617 tokens: 100 video, 16 action, one state, and 250 each for past and future tracks; 50 clean observation tokens are prepended once. Video uses factorized 3D RoPE, while action, state, and tracks use separate 1D RoPE. Tracks additionally encode view, hand, semantic role, point, block, and past/future identity. The \stagetwo{} is a 6-block bidirectional transformer ($d=768$, 12 heads, FFN $d=3072$), with cached coarse states projected from 3072 to 768 for cross-attention. All components, including the track codec, tactile VQ module, and modality embeddings, are trained from scratch end to end. Training uses $K=4$ teacher-forced blocks ($3.2$\,s); inference generates one block at a time autoregressively. Tactile dropout zeros all tactile tokens with probability $0.1$ per sample.

\begin{table}[t!]
\centering
\caption{Real-robot evaluation on the ten-task suite (scoring in
\Cref{sec:setup}). For pipette volume, the indented second row gives the
mean absolute error in $\mu$L over the engaged trials.
\emph{w/o force pred} and \emph{w/o track pred} remove future-force prediction and visual track prediction during training, respectively.}
\vspace{-0.5em}
\footnotesize
\label{tab:baselines}
\setlength{\tabcolsep}{2pt}
\begin{tabular}{@{}clccccc@{}}
\toprule
& & & & \multicolumn{3}{c@{}}{\model{} (ours)} \\
\cmidrule(l){5-7}
\# & Task & $\pi_{0.5}$ & T-Rex & \makecell{w/o\\force pred} & \makecell{w/o\\track pred} & full \\
\midrule
\multicolumn{7}{@{}l}{\emph{Binary tasks: success rate (\%) $\uparrow$}} \\
1  & Lab pipetting                 & 0    & 0    & 30   & 20        & \best{40}  \\
2  & Open test-tube cap            & 10   & 20   & 40   & \best{80} & \best{80}  \\
3  & Open microcentrifuge flip cap & 0    & 20   & 20   & 40        & \best{50}  \\
4  & Tie shoelaces                 & 0    & 0    & 0    & \best{20} & \best{20}  \\
5  & Shuffle cards                 & 0    & 10   & 0    & 20        & \best{30}  \\
6  & Multi-object grasp            & 0    & 0    & 10   & \best{40} & \best{40}  \\
7  & Use screwdriver               & 0    & 10   & 0    & \best{30} & \best{30}       \\
   & Mean                          & 1.4  & 8.6  & 14.3 & 35.7      & \best{41.4}       \\
\midrule
\multicolumn{7}{@{}l}{\emph{Continuous tasks: clean notes of 10 $\uparrow$}} \\
8  & Play harp                     & 0.5  & 2.1  & 0    & 6.5       & \best{6.7} \\
9  & Play xylophone                & 0 & 3.2 & 1.9 & 3.9      & \best{5.4}       \\
\midrule
\multicolumn{7}{@{}l}{\emph{Volume adjustment: engaged trials of 10 $\uparrow$}} \\
10 & Pipette volume                & 1    & 2    & 2    & 4         & \best{5}   \\
   & \quad mean abs.\ error ($\mu$L) $\downarrow$ & 17.0 & 14.0 & 11.0 & 12.8 & \best{6.4} \\
\bottomrule
\end{tabular}
\vspace{-2em}
\end{table}

\subsection{Training Recipe}
\label{sec:trainrecipe}
The Track-Guided Coarse Planner and the Force--Action Joint
Refiner are jointly trained by:
\begin{equation}
    \mathcal{L}_{\mathrm{FACT}}
    =
    \mathcal{L}_{\mathrm{coarse}}
    +
    \lambda_{\mathrm{refine}}\mathcal{L}_{\mathrm{refine}},
\end{equation} 
We pretrain \model{} on a mixture of 2{,}000 hours of \emph{EgoForce} human
demonstrations (\Cref{sec:egoforce}) and 100 hours of T-Rex robot
teleoperation data~\cite{trex}, sampled proportionally to hours, on 240
Alibaba T-Head PPU accelerators (Zhenwu 810E, 96\,GB HBM2e;
\Cref{tab:config}). The entire model is trained end to end under the joint
objective above: the full DiT backbone is fine-tuned from Wan2.2-TI2V-5B,
and the track codec, joint refiner, tactile VQ, and modality embeddings are
trained from scratch. Unlike a vision-only
pretraining stage, the calibrated glove forces supervise the force
pathway from the first step, so contact behavior is learned jointly with
visuomotor priors rather than grafted on afterwards.

\section{Experiments}
\label{sec:exp}

\begin{figure*}[t]
\centering
\includegraphics[width=\textwidth]{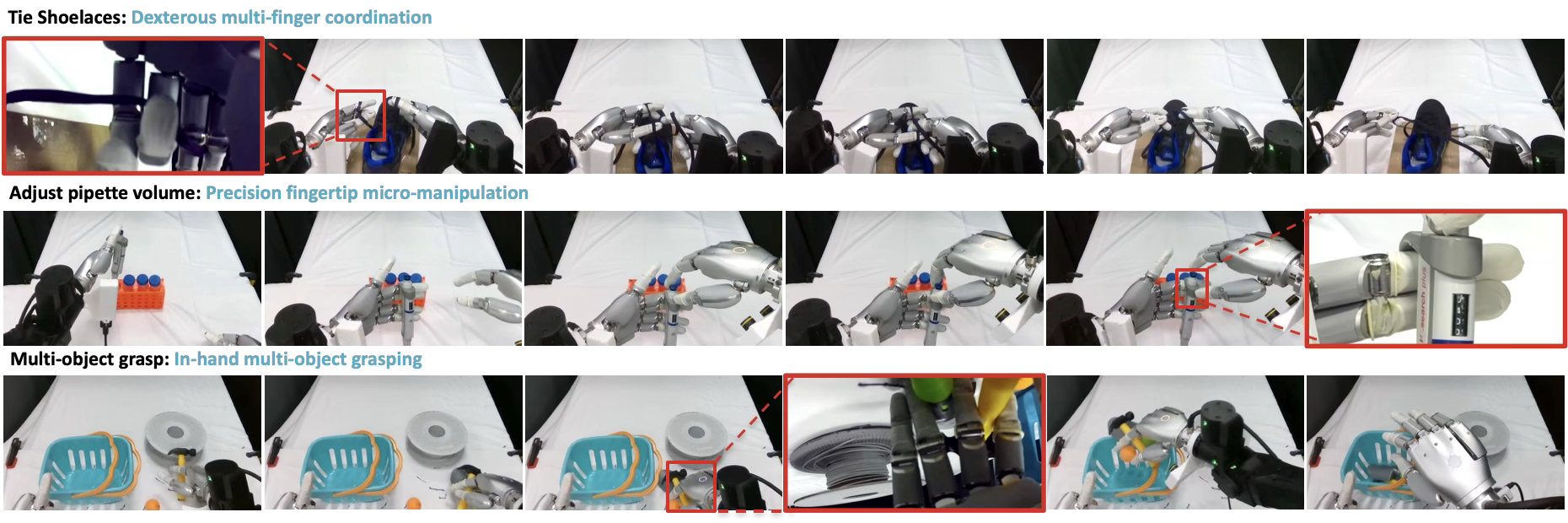}
\vspace{-1.5em}
\caption{\textbf{Qualitative examples.} \model{} executes contact-rich behaviors requiring dexterous multi-finger
coordination, precision fingertip micro-manipulation, and in-hand multi-object
grasping. 
}
\vspace{-1em}
\label{fig:rollout}
\end{figure*}

\begin{table}[t]
\centering\footnotesize
\caption{ Pretraining recipe ablation on all ten tasks.}
\vspace{-0.5em}
\label{tab:recipe}
\setlength{\tabcolsep}{2pt}
\begin{tabular}{@{}clccc@{}}
\toprule
\# & Task & \makecell{human w/\\force+robot\\(full)} & \makecell{human w/o\\force+robot} & \makecell{robot\\only} \\
\midrule
\multicolumn{5}{@{}l}{\emph{Binary tasks: success rate (\%) $\uparrow$}} \\
1  & Lab pipetting                 & \best{40}  & 20        & 0    \\
2  & Open test-tube cap            & \best{80}  & 0         & 20   \\
3  & Open microcentrifuge flip cap & \best{50}  & 10        & 0    \\
4  & Tie shoelaces                 & \best{20}  & 0         & 0    \\
5  & Shuffle cards                 & \best{30}  & 10        & 0    \\
6  & Multi-object grasp            & \best{40}  & \best{40} & 30   \\
7  & Use screwdriver               & \best{30}  & 10      & 0 \\
\midrule
\multicolumn{5}{@{}l}{\emph{Continuous tasks: clean notes of 10 $\uparrow$}} \\
8  & Play harp                     & \best{6.7} & 2.0       & 1.2  \\
9  & Play xylophone                & \best{5.4} & 2.2      & 1.0 \\
\midrule
\multicolumn{5}{@{}l}{\emph{Volume adjustment: engaged trials of 10 $\uparrow$}} \\
10 & Pipette volume                & \best{5}   & 2         & 0    \\
   & \quad mean abs.\ error ($\mu$L) $\downarrow$ & \best{6.4} & 20.5      & \tbd \\
\bottomrule
\end{tabular}
\vspace{-2em}
\end{table}

\subsection{Experimental Setup}
\label{sec:setup}

\paragraph{Robot platform and data collection}
We evaluate \model{} on a ten-task real-robot suite.
All policies are evaluated on a bimanual YAM station with two Sharpa Wave hands (22 joints, five-fingertip 6-axis force sensing), equipped with a RealSense D435 ego-view camera and two ZED Mini wrist cameras. For each evaluation task we collect 10 robot teleoperation trajectories using a Manus-glove and wrist-tracker interface and 90 human demonstrations collected with a Wuji tactile glove and wrist-tracker interface, resulting in a 100-trajectory mixture dataset per task.


\paragraph{Skill-specific fine-tuning}


We fine-tune the pretrained \model{} on the 100-trajectory dataset for each task on 8 GPUs. This preserves the force-reactive behaviors acquired in pretraining while adapting to specific tasks.\looseness=-1

\paragraph{Baselines and ablations}
We compare with two baselines, $\pi_{0.5}$~\cite{black2025pi05} and T-Rex~\cite{trex}, each fine-tuned from its public base checkpoint on the same dataset. We additionally compare two \model{} ablations: \emph{w/o force pred} and \emph{w/o track pred}, which remove future-force prediction and visual track prediction during training, respectively. 

\paragraph{Evaluation protocol and metrics}
The suite covers ten contact-rich tasks in three scoring classes.
\emph{Binary tasks} (seven: lab pipetting, test-tube cap opening,
microcentrifuge flip-cap opening, shoelace tying, card shuffling,
multi-object grasping, screwdriver use) score success over ten rollouts.
\emph{Continuous tasks} (two) pluck each of ten indicated harp strings or
strike each of ten indicated xylophone bars and score the number of notes
that sound cleanly.
\emph{Volume adjustment} starts the pipette dial at 420\,$\mu$L with a
400\,$\mu$L target; we report the number of trials (of ten) that
\emph{engage} the dial and, in an indented second row, the mean absolute
error of the final reading over engaged trials. Reporting both prevents
rare engagement from appearing precise; a never-engaged trial is a failure
with no reading. All methods share the same scenes and objects across
rollouts, with object placement randomized per rollout
(\Cref{fig:randomness}).\looseness=-1


\begin{figure*}[h!]
\centering
\includegraphics[width=\textwidth]{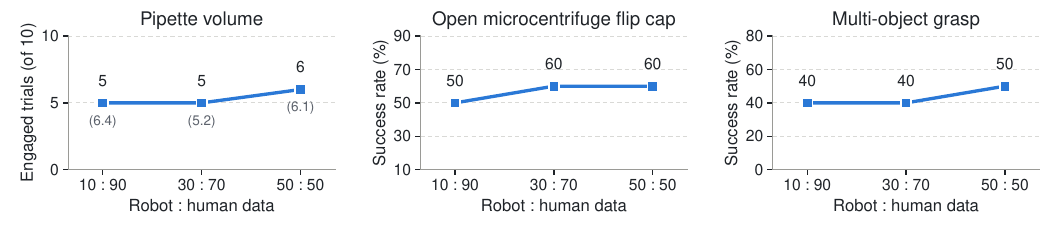}
\vspace{-2.5em}
\caption{\textbf{Robot-to-human mixture in fine-tuning.} Performance
versus the robot-to-human ratio of the 100-trajectory per-task fine-tuning
mixture (10:90, 30:70, 50:50).}
\vspace{-1em}
\label{fig:ratio}
\end{figure*}

\begin{table}[!t]
\centering\footnotesize
\caption{Contribution of human fine-tuning data on three dexterous
tasks: ten robot trajectories alone (10:0) versus the 10:90
robot-to-human mixture, scored as in \Cref{tab:baselines}.}
\vspace{-0.5em}
\label{tab:humandata}
\setlength{\tabcolsep}{3pt}
\begin{tabular}{@{}clcc@{}}
\toprule
& & \multicolumn{2}{c@{}}{robot : human data} \\
\cmidrule(l){3-4}
\# & Task & 10 : 0 & 10 : 90 \\
\midrule
\multicolumn{4}{@{}l}{\emph{Binary tasks: success rate (\%) $\uparrow$}} \\
1  & Tie shoelaces & 10    & \best{20} \\
2  & Shuffle cards & 10 & \best{30} \\
\midrule
\multicolumn{4}{@{}l}{\emph{Volume adjustment: engaged trials of 10 $\uparrow$}} \\
3  & Pipette volume                & 4    & \best{5}   \\
   & \quad mean abs.\ error ($\mu$L) $\downarrow$ & 11.2 & \best{6.4} \\
\bottomrule
\end{tabular}
\vspace{-2em}
\end{table}

\subsection{Main Results}
\label{sec:baselines}
\Cref{tab:baselines} shows that \model{} consistently outperforms the
baselines across the ten-task suite. On representative binary tasks, it
achieves 40/80/40\% success on pipetting, test-tube cap opening, and
multi-object grasping, compared with 0/20/0\% for T-Rex and 0/10/0\% for
$\pi_{0.5}$. It also sounds 6.7 of 10 harp notes cleanly, compared with
2.1 for T-Rex and 0.5 for $\pi_{0.5}$. The difference is especially pronounced
on force-critical tasks: \model{} engages the pipette-volume dial in 5 of 10 trials with 6.4\,$\mu$L mean error, compared with 2 engagements and
14\,$\mu$L error for T-Rex. Among the five successful engagements, four
finish within 4\,$\mu$L of the target, suggesting that failures arise
primarily from engaging the dial rather than adjusting it once engaged.
Together with the harp results, these results demonstrate \model{}'s ability to translate force-rich human pretraining and joint continuous force--action denoising into precise, contact-sensitive robot behavior.
\Cref{fig:rollout} illustrates this behavior on three representative
tasks: shoelace tying, where both hands pinch, cross, and pull the laces with
coordinated multi-finger regrasps; pipette-volume adjustment, where the
fingertips turn the dial in small increments while the rest of the hand
holds the pipette steady; and multi-object grasping, where the hand secures
a second object without releasing the first. In each case success is decided
by fingertip contact rather than gross arm motion.\looseness=-1



\begin{figure}[t]
\centering
\includegraphics[width=\linewidth]{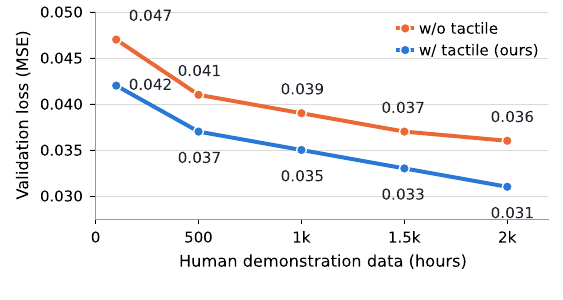}
\vspace{-2.5em}
\caption{\textbf{Scaling with human data.} Validation loss (MSE)
versus hours of human demonstration data (100, 500, 1k, 1.5k, and 2k hours on a
linear axis), mixed with robot teleoperation data, with and without tactile supervision.
Tactile supervision lowers the loss at every scale and the gap persists as data grows.}
\label{fig:scaling}
\vspace{-2em}
\end{figure}

\subsection{Ablation Studies}
\label{sec:ablations}

\paragraph{Force and track prediction}
\Cref{tab:baselines} shows the effectiveness of future-force and visual track prediction during training. Removing force prediction causes substantial degradation on contact-sensitive tasks, including test-tube cap opening ($80\to40\%$),
harp playing ($6.7\to0$ notes), and pipette-volume adjustment
($5/10\to2/10$ engagements). Removing track prediction produces smaller but consistent
drops on tasks requiring precise motion, including pipetting ($40\to20\%$),
card shuffling ($30\to20\%$), and xylophone playing ($5.4\to3.9$ notes).
Together, these results show force prediction primarily supports
contact-rich execution, while tracks improve precise motion.\looseness=-1

\paragraph{Force supervision in human pretraining data}
To isolate the contribution of force-supervised human data, we compare three
pretraining recipes across the ten tasks (\Cref{tab:recipe}): human
demonstrations with glove-measured force mixed with robot teleoperation,
the same mixture without human force supervision, and robot-only training.
Removing force supervision causes the largest drops on contact-sensitive
tasks: test-tube cap opening falls from 80\% to 0\%, pipetting from 40\%
to 20\%, pipette-dial engagement from 5/10 to 2/10 with mean error
increasing from 6.4 to 20.5\,$\mu$L, and xylophone performance from 5.4
to 2.2 clean notes. In contrast, multi-object grasping remains unchanged
at 40\%. Removing human data entirely also degrades performance, yielding
0\% on pipetting and flip-cap opening, 1.2 clean harp notes, and no
successful dial engagement. These results show that \model{} benefits substantially from human demonstrations during pretraining, with force supervision crucial to translating this experience into strong performance on robot dexterous manipulation tasks.

\paragraph{Scaling with pretraining human data}
\label{sec:scaling}
To test how downstream robot policy performance scales with pretraining human data, we train \model{} with
100, 500, 1k, 1.and 2k hours of human demonstrations, each mixed with the same
robot teleoperation data, and compare training with and without tactile
supervision. \Cref{fig:scaling} shows that validation loss decreases
log-linearly with human data in both settings, while tactile supervision
provides a consistent 10--14\% improvement at every scale
($0.047{\to}0.042$ at 100\,h and $0.036{\to}0.031$ at 2k\,h).
The benefit persists as data scales: from 1k to 2k hours, validation loss
decreases by 11\% with tactile supervision, compared with 8\% without it.
All models are trained on 240 Alibaba T-Head PPUs.

\paragraph{Human data in fine-tuning}
\Cref{tab:humandata} isolates the contribution of human data on three
dexterous tasks by comparing fine-tuning with ten robot trajectories alone
(10:0) against the full 10:90 robot--human mixture. Adding the 90 human
demonstrations improves shoelace tying from 10\% to 20\%, card shuffling from 10\% to 30\%, and
pipette-dial engagement from 4/10 to 5/10 while reducing mean error from
11.2 to 6.4\,$\mu$L. These gains demonstrate \model{}'s ability to transfer manipulation knowledge from human demonstrations into improved robot task completion and precision, making effective use of human experience when robot data is limited.

\paragraph{Human-demonstration efficiency in fine-tuning}
\Cref{fig:ratio} examines how performance changes with the robot-to-human data ratio in the 100-trajectory per-task fine-tuning mixture, using robot-to-human ratios of 10:90, 30:70, and 50:50.
Increasing robot data yields modest gains: flip-cap opening improves from
50\% to 60\%, multi-object grasping from 40\% to 50\%, and pipette-dial
engagement from 5/10 to 6/10. Pipette-volume error remains similar across
the three mixtures ($6.4$, $5.2$, and $6.1$\,$\mu$L). These results suggest that \model{} learns from human demonstrations nearly
as effectively as from robot demonstrations for fine-tuning.

\section{Conclusion}
\label{sec:conclusion}
We present \model{}, a dexterous robot foundation model with force-aware
human pretraining. By combining large-scale force-rich human experience
with robot demonstrations, \model{} couples prospective visual planning with
force-aware action refinement: future video and visual tracks guide coarse
motion, while predicted future force shapes fine-grained execution. Across ten
dexterous tasks, our experiments demonstrate the complementary roles of tracks
and force prediction and show that human demonstrations can substantially
reduce the amount of task-specific robot data required for fine-tuning.

\bibliographystyle{IEEEtranN}
\bibliography{references}

\begin{thebibliography}{38}
\providecommand{\natexlab}[1]{#1}
\providecommand{\url}[1]{#1}
\csname url@samestyle\endcsname
\providecommand{\newblock}{\relax}
\providecommand{\bibinfo}[2]{#2}
\providecommand{\BIBentrySTDinterwordspacing}{\spaceskip=0pt\relax}
\providecommand{\BIBentryALTinterwordstretchfactor}{4}
\providecommand{\BIBentryALTinterwordspacing}{\spaceskip=\fontdimen2\font plus
\BIBentryALTinterwordstretchfactor\fontdimen3\font minus \fontdimen4\font\relax}
\providecommand{\BIBforeignlanguage}[2]{{%
\expandafter\ifx\csname l@#1\endcsname\relax
\typeout{** WARNING: IEEEtranN.bst: No hyphenation pattern has been}%
\typeout{** loaded for the language `#1'. Using the pattern for}%
\typeout{** the default language instead.}%
\else
\language=\csname l@#1\endcsname
\fi
#2}}
\providecommand{\BIBdecl}{\relax}
\BIBdecl

\bibitem[Todorov and Jordan(2002)]{todorov2002optimal}
E.~Todorov and M.~I. Jordan, ``Optimal feedback control as a theory of motor coordination,'' \emph{Nature Neuroscience}, vol.~5, no.~11, pp. 1226--1235, 2002.

\bibitem[Scott(2004)]{scott2004optimal}
S.~H. Scott, ``Optimal feedback control and the neural basis of volitional motor control,'' \emph{Nature Reviews Neuroscience}, vol.~5, no.~7, pp. 532--545, 2004.

\bibitem[Grauman et~al.(2022)]{grauman2022ego4d}
K.~Grauman \emph{et~al.}, ``{Ego4D}: Around the world in 3,000 hours of egocentric video,'' in \emph{CVPR}, 2022.

\bibitem[Damen et~al.(2018)]{damen2018epic}
D.~Damen \emph{et~al.}, ``Scaling egocentric vision: The {EPIC-KITCHENS} dataset,'' in \emph{ECCV}, 2018.

\bibitem[Nair et~al.(2022)]{nair2022r3m}
S.~Nair \emph{et~al.}, ``{R3M}: A universal visual representation for robot manipulation,'' \emph{arXiv preprint arXiv:2203.12601}, 2022.

\bibitem[Wang et~al.(2023)]{wang2023mimicplay}
C.~Wang \emph{et~al.}, ``{MimicPlay}: Long-horizon imitation learning by watching human play,'' \emph{Conference on Robot Learning (CoRL)}, 2023.

\bibitem[Chi et~al.(2024)]{chi2024umi}
C.~Chi \emph{et~al.}, ``Universal manipulation interface: In-the-wild robot teaching without in-the-wild robots,'' \emph{Robotics: Science and Systems (RSS)}, 2024.

\bibitem[Wang et~al.(2024)]{wang2024dexcap}
C.~Wang \emph{et~al.}, ``{DexCap}: Scalable and portable mocap data collection system for dexterous manipulation,'' \emph{arXiv preprint arXiv:2403.07788}, 2024.

\bibitem[Kareer et~al.(2024)]{kareer2024egomimic}
S.~Kareer \emph{et~al.}, ``{EgoMimic}: Scaling imitation learning via egocentric video,'' \emph{arXiv preprint arXiv:2410.24221}, 2024.

\bibitem[Hoque et~al.(2025)]{hoque2025egodex}
R.~Hoque \emph{et~al.}, ``{EgoDex}: Learning dexterous manipulation from large-scale egocentric video,'' \emph{arXiv preprint arXiv:2505.11709}, 2025.

\bibitem[Niu et~al.(2026)]{trex}
D.~Niu \emph{et~al.}, ``{T-Rex}: Tactile-reactive dexterous manipulation,'' 2026.

\bibitem[Yuan et~al.(2017)Yuan, Dong, and Adelson]{yuan2017gelsight}
W.~Yuan, S.~Dong, and E.~H. Adelson, ``{GelSight}: High-resolution robot tactile sensors for estimating geometry and force,'' \emph{Sensors}, 2017.

\bibitem[Wu et~al.(2024)]{wu2024threedvitac}
B.~Wu \emph{et~al.}, ``{3D-ViTac}: Learning fine-grained manipulation with visuo-tactile sensing,'' \emph{arXiv preprint arXiv:2410.24091}, 2024.

\bibitem[Higuera et~al.(2024)]{higuera2024sparsh}
C.~Higuera \emph{et~al.}, ``Sparsh: Self-supervised touch representations for vision-based tactile sensing,'' \emph{Conference on Robot Learning (CoRL)}, 2024.

\bibitem[Guzey et~al.(2023)]{guzey2023tdex}
I.~Guzey \emph{et~al.}, ``Dexterity from touch: Self-supervised pre-training of tactile representations,'' \emph{CoRL}, 2023.

\bibitem[Qi et~al.(2023)]{qi2023general}
H.~Qi \emph{et~al.}, ``General in-hand object rotation with vision and touch,'' \emph{CoRL}, 2023.

\bibitem[Heng et~al.(2025)]{heng2025vitacformer}
L.~Heng \emph{et~al.}, ``{ViTacFormer}: Learning cross-modal representation for visuo-tactile dexterous manipulation,'' \emph{arXiv preprint}, 2025.

\bibitem[Huang et~al.(2025)]{huang2025tactilevla}
J.~Huang \emph{et~al.}, ``{Tactile-VLA}: Unlocking vision-language-action model's physical knowledge for tactile generalization,'' \emph{arXiv preprint arXiv:2507.09160}, 2025.

\bibitem[Hu et~al.(2024)]{hu2024forcemimic}
W.~Hu \emph{et~al.}, ``{ForceMimic}: Force-centric imitation learning with force-motion capture system for contact-rich manipulation,'' \emph{arXiv preprint arXiv:2410.07554}, 2024.

\bibitem[Li et~al.(2024)]{li2024forceaware}
H.~Li \emph{et~al.}, ``Force-aware imitation learning for contact-rich manipulation,'' \emph{arXiv preprint}, 2024.

\bibitem[Belkhale et~al.(2023)Belkhale, Cui, and Sadigh]{belkhale2023hydra}
S.~Belkhale, Y.~Cui, and D.~Sadigh, ``{HYDRA}: Hybrid robot actions for imitation learning,'' \emph{CoRL}, 2023.

\bibitem[Belkhale et~al.(2024)]{belkhale2024rth}
S.~Belkhale \emph{et~al.}, ``{RT-H}: Action hierarchies using language,'' \emph{arXiv preprint arXiv:2403.01823}, 2024.

\bibitem[Black et~al.(2025)]{black2025pi06}
K.~Black \emph{et~al.}, ``$\pi^*_{0.6}$: A {VLA} that learns from experience,'' \emph{arXiv preprint}, 2025.

\bibitem[Xue et~al.(2025)]{xue2025rdp}
H.~Xue \emph{et~al.}, ``Reactive diffusion policy: Slow-fast visual-tactile policy learning for contact-rich manipulation,'' \emph{arXiv preprint arXiv:2503.02881}, 2025.

\bibitem[van~den Oord et~al.(2017)van~den Oord, Vinyals, and Kavukcuoglu]{vandenoord2017vqvae}
A.~van~den Oord, O.~Vinyals, and K.~Kavukcuoglu, ``Neural discrete representation learning,'' \emph{NeurIPS}, 2017.

\bibitem[Wu et~al.(2023)]{wu2023gr1}
H.~Wu \emph{et~al.}, ``Unleashing large-scale video generative pre-training for visual robot manipulation,'' \emph{arXiv preprint arXiv:2312.13139}, 2023.

\bibitem[Du et~al.(2023)]{du2023unipi}
Y.~Du \emph{et~al.}, ``Learning universal policies via text-guided video generation,'' \emph{NeurIPS}, 2023.

\bibitem[Yang et~al.(2024)]{yang2024unisim}
M.~Yang \emph{et~al.}, ``Learning interactive real-world simulators,'' \emph{ICLR}, 2024.

\bibitem[Agarwal et~al.(2025)]{agarwal2025cosmos}
N.~Agarwal \emph{et~al.}, ``Cosmos world foundation model platform for physical {AI},'' \emph{arXiv:2501.03575}, 2025.

\bibitem[Wen et~al.(2024)]{wen2023atm}
C.~Wen \emph{et~al.}, ``Any-point trajectory modeling for policy learning,'' \emph{RSS}, 2024.

\bibitem[Bharadhwaj et~al.(2024)]{bharadhwaj2024track2act}
H.~Bharadhwaj \emph{et~al.}, ``{Track2Act}: Predicting point tracks from internet videos enables generalizable robot manipulation,'' \emph{ECCV}, 2024.

\bibitem[Xu et~al.(2024)]{xu2024flow}
M.~Xu \emph{et~al.}, ``Flow as the cross-domain manipulation interface,'' \emph{CoRL}, 2024.

\bibitem[Cao et~al.(2026)]{cao2026tractbridgingrobotcontrol}
Z.~Cao \emph{et~al.}, ``{TrAct}: Bridging robot control and visual prediction with visual tracks,'' arXiv preprint arXiv:2608.24101, 2026.

\bibitem[Karaev et~al.(2024)]{karaev2024cotracker}
N.~Karaev \emph{et~al.}, ``{CoTracker}: It is better to track together,'' \emph{ECCV}, 2024.

\bibitem[{Wan Team}(2025)]{wan2025}
{Wan Team}, ``{Wan}: Open and advanced large-scale video generative models,'' \emph{arXiv:2503.20314}, 2025.

\bibitem[Chung et~al.(2023)]{chung2023unimaxfairereffectivelanguage}
H.~W. Chung \emph{et~al.}, ``{UniMax}: Fairer and more effective language sampling for large-scale multilingual pretraining,'' arXiv preprint arXiv:2304.09151, 2023.

\bibitem[Radford et~al.(2021)]{radford2021learningtransferablevisualmodels}
A.~Radford \emph{et~al.}, ``Learning transferable visual models from natural language supervision,'' ICML, 2021.

\bibitem[{Physical Intelligence} et~al.(2025){Physical Intelligence}, Black, et~al.]{black2025pi05}
{Physical Intelligence}, K.~Black \emph{et~al.}, ``$\pi_{0.5}$: A vision-language-action model with open-world generalization,'' \emph{arXiv preprint arXiv:2504.16054}, 2025.

\end{thebibliography}

\section*{Acknowledgements}
The 2000 hours of vision-tactile sensing data used for model training in this
work were collected with the FusionX multimodal data collection system
(TouchTronix Robotics, NY, USA). The authors thank Dr.\ Jun Liu and Dr.\
Yixiang Gao of TouchTronix Robotics for fruitful scientific discussions on
this manuscript.

The authors thank Sharpa and Xdof for the donation of the dexterous hands and robot arms used in the experiments.

\appendices

\section{Model and Training Details}
\label{app:model}

\begin{table}[H]
\centering\footnotesize
\caption{Model and training configuration of \model{}.}
\label{tab:config}
\setlength{\tabcolsep}{3pt}\renewcommand{\arraystretch}{1.05}
\begin{tabular}{@{}p{3.3cm}p{5.1cm}@{}}
\toprule
\multicolumn{2}{@{}l}{\emph{\stageone{} (Flow Matching)}} \\
Backbone & Wan2.2-TI2V-5B video DiT \\
Hidden dimension & 3072 \\
Number of layers & 30 \\
Max sequence length & 2{,}518 tokens \\
Action dimension & 62 \\
Action chunk & 16 \\
Timestep sampling & $\mathrm{Beta}(1.5, 1.0)$ \\
Num inference timesteps & 6 ($\tau\!:1{\to}0.4$) \\
Parameters & 5.05B \\
\midrule
\multicolumn{2}{@{}l}{\emph{\stagetwo{} (Flow Matching)}} \\
Hidden dimension & 768 \\
Number of layers & 6 \\
FFN intermediate size & 3072 \\
Action dimension & 62 \\
Action chunk & 16 \\
Num inference timesteps & 4 \\
Parameters & 65M \\
\midrule
\multicolumn{2}{@{}l}{\emph{Training configuration}} \\
Optimizer & AdamW \\
Peak learning rate & $1{\times}10^{-5}$ \\
Min learning rate & 0 \\
LR scheduler & cosine \\
Warmup ratio & 0.05 \\
Weight decay & $10^{-5}$ \\
Gradient clipping & 1.0 \\
GPU type & Alibaba T-Head PPU (Zhenwu 810E, 96\,GB HBM2e) \\
Number of GPUs & 240 \\o
DeepSpeed ZeRO stage & 2 \\
Per device batch size & 16 \\
Gradient accumulation steps & 1 \\
Mixed precision training & bf16 \\
Tactile dropout & 0.1 \\
\bottomrule
\end{tabular}
\end{table}

\Cref{tab:config} lists the configuration of both stages and the training
setup. The \stageone{} is the Wan2.2-TI2V-5B diffusion transformer fine-tuned
in full; the action, state, track, and image projections, the modality-type
embedding, and the entire \stagetwo{} are trained from scratch, while the
VAE, the umT5-XXL text encoder, and the CLIP image encoder stay frozen.

\paragraph{Visual tracks}
Tracks are extracted offline: SAM2 segments the hands in the first frame and
CoTracker follows 250 points per block through the episode at
$180{\times}320$ (50 points per hand in the head view; 50 hand points plus a
$5{\times}5$ background grid in each wrist view), normalized to $[0,1]$, with
a 16-frame past horizon and a 16-step future horizon at 20\,Hz. The track
encoder flattens $[x,y,\text{vis},\Delta x,\Delta y]$ over 16 frames through
$\mathrm{Linear}(80{\to}3072)\to\mathrm{SiLU}\to\mathrm{Linear}$ with view,
hand, role, point, block, and past/future embeddings, and the decoder maps
each future-track token to 16 two-dimensional velocities. Track losses are
visibility-masked and weighted by $1+4\ell$, where $\ell$ is the point's
path length.

\paragraph{Data pipeline}
Demonstrations are stored in LeRobot-v2 format at 30\,fps and resampled by
nearest timestamp to 20\,Hz for actions and tracks, 10\,Hz for video, and
5\,Hz for tactile force. The 62-D action concatenates, per arm, the
end-effector position and 6-D rotation relative to the chunk's starting pose
and 22 absolute hand joints; states, actions, and forces are normalized by
their 1st/99th percentiles to $[-1,1]$. The three camera views are resized
to $160{\times}320$ and tiled $2{\times}2$ before VAE encoding, so that each
latent frame yields 50 tokens.

\paragraph{Inference}
The coarse stage runs six Euler steps from $\tau{=}1$ to $0.4$ and one
further forward pass at $\tau{=}0.4$ that reads out video and tracks and
produces the hidden states cached as the refiner's cross-attention memory;
the refiner then runs four joint Euler steps. No classifier-free guidance is
used, and the resulting 16-step chunk is executed at 20\,Hz.

\section{Force--Action Joint Refiner}
\label{app:force}

\paragraph{Tactile input}
Each hand provides five fingertip 6-axis wrenches $(F_x,F_y,F_z,M_x,M_y,M_z)$,
giving a 60-D force observation per time step, normalized per channel by its
1st/99th percentiles. At each 5\,Hz refresh slot the refiner receives the force
history of the last 16 samples (3.2\,s); missing samples at the start of an
episode are zeroed and masked, and the current force is the last sample of the
window.

\paragraph{Force-history quantization}
The history of every fingertip is encoded by a 1D convolutional encoder shared
across both hands: kernel-5 convolutions with GroupNorm and GELU, 128
channels, two stride-2 blocks, and a 256-D bottleneck that is mean-pooled over
time into one 256-D code per fingertip. The codebook has 64 entries of 256
dimensions, updated by exponential moving average (decay 0.99) with dead codes
re-seeded every 200 steps. The VQ loss is the reconstruction MSE plus a
commitment term with weight 0.25, added to the total objective with weight 1;
learned hand and finger embeddings are added to the resulting ten history
tokens.

\paragraph{Token layout}
A refiner instance operates on 76 tokens: 16 action tokens
($\mathrm{Linear}(64{\to}768)$ plus step-position, slot, type, and
refresh-offset embeddings), 10 current-force tokens (a $6{\to}768{\to}768$ MLP
plus finger and type embeddings), 10 VQ-history tokens
($\mathrm{Linear}(256{\to}768)$), and 40 future-force tokens for the four
5\,Hz slots of the chunk, which reuse the force MLP with finger and slot
embeddings. Every block applies per-token AdaLN modulation (sinusoidal
embedding of $1000\tau$ through an MLP to six 768-D vectors, plus a learned
per-block bias), full bidirectional self-attention over all 76 tokens,
cross-attention to the coarse memory, and an AdaLN-modulated feed-forward
network. The action head is a LayerNorm followed by
$\mathrm{Linear}(768{\to}64)$ and the force head a LayerNorm followed by
$\mathrm{Linear}(768{\to}6)$; both predict flow-matching velocities.

\paragraph{Coarse memory}
The memory consists of the planner's hidden states for the observation and
action tokens of the current block and up to four preceding blocks at
$\tau{=}0.4$ (116 tokens for a single block), projected once through
$\mathrm{Linear}(3072{\to}768)$ and a LayerNorm and reused across all refiner
steps. During training the memory is produced by an extra forward pass of the
planner at exactly $\tau{=}0.4$, without gradient, using the same suffix noise
draw as the refiner so that training matches the cascaded inference state.

\paragraph{Training}
For every training sample the refiner is instantiated for each of the four
blocks and each refresh offset $o\in\{0,4,8,12\}$, i.e.\ 16 instances per
sample. The action prefix $a_{[0:o)}$ is clean, and the suffix receives
$\tau_a=0.4\,(0.999\,\mathrm{Beta}(1.5,1)+0.001)$; force slots at or before the
offset are clean and later slots receive $\tau_f\!\sim\!\mathrm{Beta}(1.5,1)$.
The future-force targets are the normalized measured forces at the four slot
times. The loss is the flow-matching error on the action suffix plus 0.5 times
the error on the unobserved force slots, plus the VQ loss. Tactile dropout
zeroes the current-force and history tokens with probability 0.1 per sample
so the refiner cannot rely on touch alone.

\paragraph{Sampling and execution}
At inference the refiner takes four joint Euler steps with $\tau_a$ on a
linear grid from 0.4 to 0 and $\tau_f$ from 1 to 0, updating only the action
suffix and the unobserved force slots while the executed prefix and observed
slots stay fixed; observed slots are filled with the current measured force.
The refined chunk is executed at 20\,Hz. The coarse planner runs once per
chunk; at each subsequent 5\,Hz tactile refresh (offsets 4, 8, and 12 of the
executing chunk) the refiner alone is re-run against the cached coarse
memory, with the executed prefix and the newly measured force slots clamped,
so the remaining suffix is re-refined with the latest contact information at
a fraction of the planner's cost. The predicted future-force trajectory is
recorded for analysis and is not used as a control reference.

\section{Hardware and Data Collection}
\label{app:hardware}

\begin{figure*}[t]
\centering
\includegraphics[width=\textwidth,trim=0 119bp 0 0,clip]{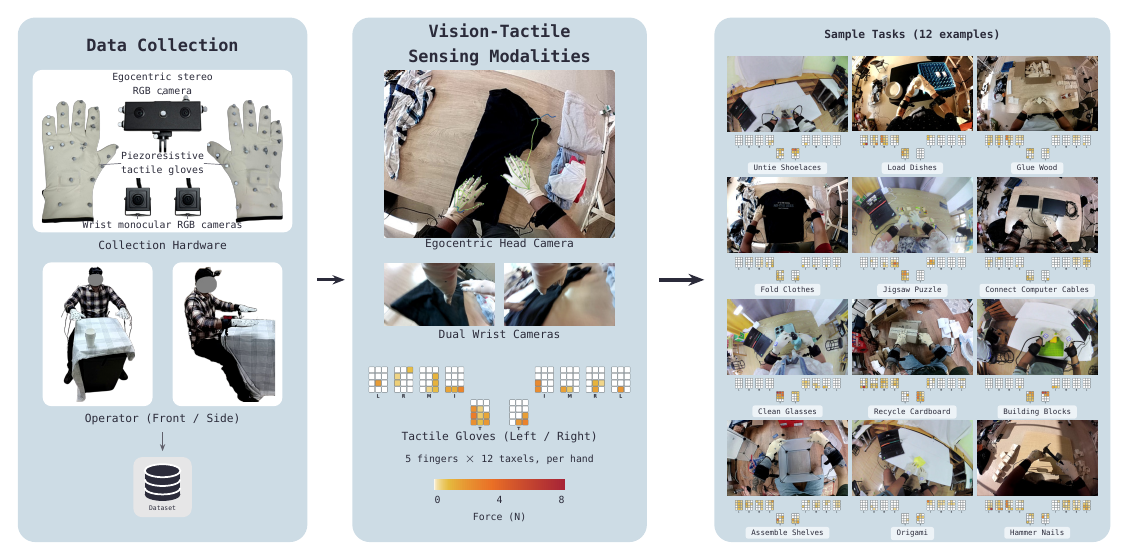}
\caption{\textbf{EgoForce collection system and data.} \textsc{Left}:
collection hardware (egocentric stereo RGB camera, wrist RGB cameras, and
piezoresistive tactile gloves) and front and side views of an operator
wearing the system. \textsc{Middle}: the FusionX multimodal collection
system (TouchTronix Robotics, NY), illustrated with synchronized head and
wrist views and tactile readings from a shirt-folding recording; the head
view includes estimated hand poses and image-space hand trajectories, which
are not motion-capture ground truth. \textsc{Right}: twelve examples from the
collection environments, each showing the egocentric view and the tactile
maps of the left and right hand (finger labels T, I, M, R, L for thumb,
index, middle, ring, and little finger). Heatmaps show per-taxel forces in
newtons on a shared 0--8\,N scale with square-root color mapping.}
\label{fig:egoforce_overview}
\end{figure*}

\paragraph{EgoForce collection (pretraining data)}
EgoForce is collected with the FusionX multimodal collection system
(\Cref{fig:egoforce_overview}). The operator wears a head-mounted stereo
camera (Arducam AR0234 global-shutter pair, two $1920{\times}1200$ RGB
imagers with a 60\,mm baseline and factory-calibrated intrinsics and
extrinsics, 30\,fps, with a 200\,Hz IMU), a monocular $1920{\times}1080$
RGB camera at 30\,fps with an IMU on each wrist, and a pair of TouchTronix
tactile gloves. Each glove carries a piezoresistive conductive-textile array
with 12 taxels on every fingertip ($5{\times}12$ per hand) and 60 taxels on
the palm, a bend sensor per finger, and an orientation IMU, all sampled at
about 200\,Hz; every sample stores the raw 8-bit A/D reading and the
calibrated force in newtons of each taxel, the five bend values, and the
hand orientation quaternion. Before a session, a per-user calibration
records the resting A/D offset of every taxel and the minimum and maximum
bend of every finger, which are subtracted and normalized at load time. All
streams are time-stamped on a common clock and stored in one MCAP file per
episode, so that any frame of the three videos can be paired with the
tactile samples of the same instant. Taxel readings are converted to
physical force with the pneumatic calibration rig of \Cref{sec:egoforce}: a
uniformly pressurized chamber and a compliant bladder conform to the glove
surface while a reference sensor records the applied pressure, giving a
per-taxel pressure--A/D curve and the force $F_i{=}P_iA_i$ from the taxel's
effective area.

\paragraph{Human action and force labels}
Hand motion is not measured by motion capture but estimated from the head
camera: ViTPose hand keypoints and HaMeR MANO hand meshes are fit to every
frame of the left stereo view, temporally smoothed, and anchored to metric
scale with the stereo depth, which yields the wrist pose and the finger
joint angles of both hands in the head-camera frame. The glove's bend
sensors and orientation IMU refine the finger flexion and hand orientation
where the camera view of the hand is occluded. The finger joint angles are
retargeted to the 22 joints of the Sharpa Wave hand and the wrist pose is
expressed relative to the start of each action chunk, matching the
delta-base convention of the robot action space (\Cref{app:model}). For
force supervision, the 12 calibrated taxel forces of each fingertip are
reduced to the 6-axis fingertip wrench used by the model: their sum gives
the normal force, their force-weighted centroid gives the center of
pressure, from which the two moments about the fingertip frame follow, and
the tangential components and twisting moment, which a pressure array
cannot observe, are set to zero. The resulting per-fingertip wrenches are
resampled to the model's 5\,Hz force slots and normalized with the same
percentile statistics as the robot wrenches.

\begin{figure*}[t]
\centering
\includegraphics[width=0.86\textwidth]{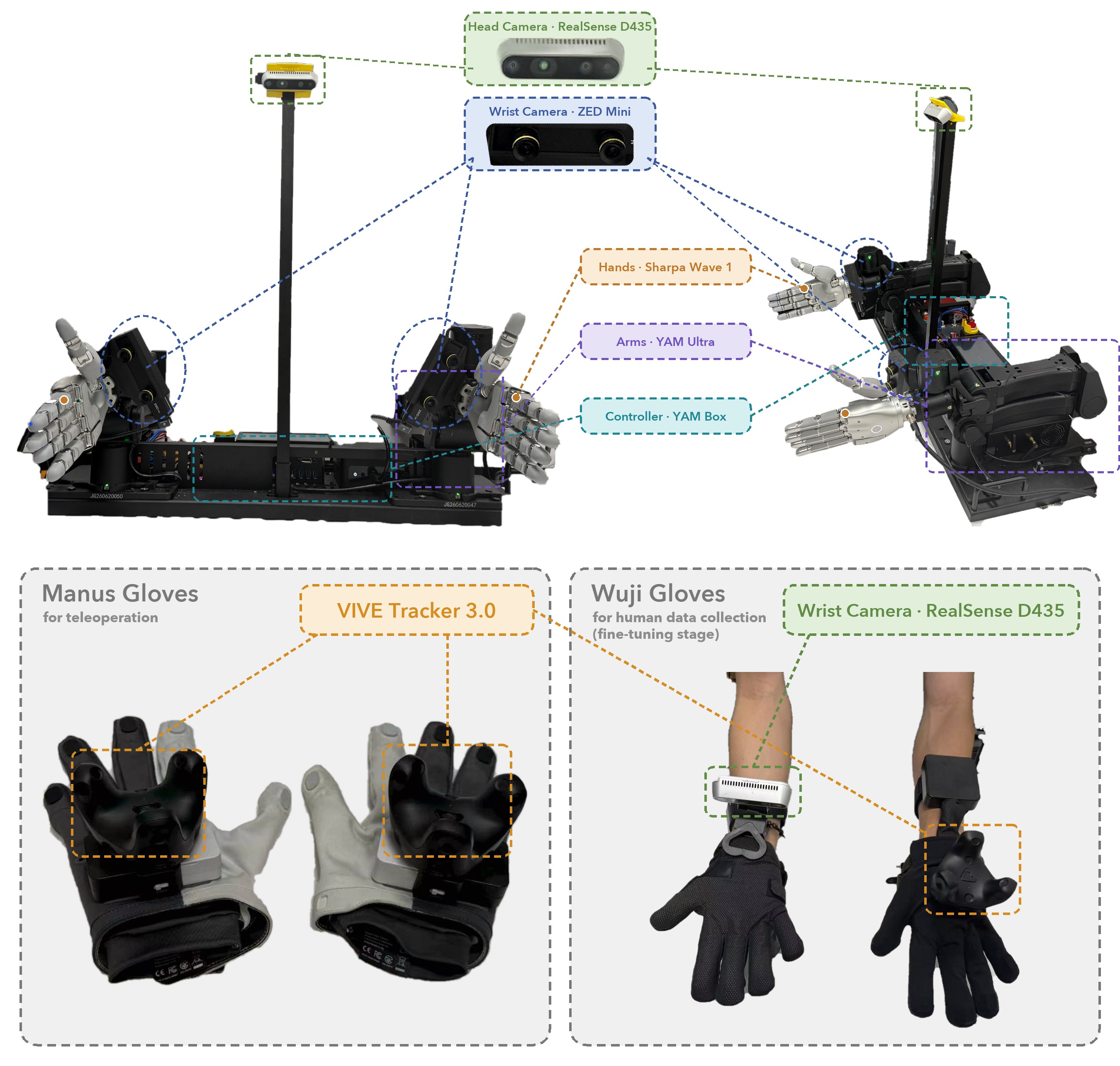}
\caption{\textbf{Real-world setup and data-collection interfaces.} Top: the
bimanual station, with two I2RT YAM Ultra arms driven by a YAM Box
controller, a Sharpa Wave 1 hand on each arm, a RealSense D435 head camera,
and a ZED Mini camera on each wrist. Bottom left: the teleoperation
interface, Manus gloves with a VIVE Tracker 3.0 on each wrist. Bottom right:
the human demonstration interface used for the fine-tuning data, Wuji
tactile gloves with a RealSense D435 wrist camera and a VIVE Tracker 3.0 on
each wrist.}
\label{fig:hardware}
\end{figure*}

\paragraph{Robot platform}
All rollouts and task-specific robot demonstrations use the bimanual station
in \Cref{fig:hardware}. Two I2RT YAM Ultra 6-DoF arms are mounted on a
common base and driven by a YAM Box controller, which also carries the
mast of the head camera. Each arm ends in a Sharpa Wave 1 dexterous hand
with 22 joints, whose five fingertips each measure a 6-axis force/torque
wrench, delivered by the hand's on-board inference at about 30\,Hz. A
RealSense D435 on the mast provides the head view and a ZED Mini on each
wrist provides the two wrist views, all at $1280{\times}720$ and 30\,fps.
The policy observes the three views, the robot state, and the ten fingertip
wrenches, and commands the 62-D action of \Cref{app:model}: a delta-base
end-effector pose per arm, resolved to joint targets by inverse kinematics
on the YAM model and tracked by the arms' built-in joint controllers, and
22 absolute joint targets per hand.

\paragraph{Teleoperation}
Robot demonstrations are collected by an operator wearing Manus gloves with
a VIVE Tracker 3.0 strapped to each wrist (\Cref{fig:hardware}, bottom
left). The glove's finger motion is retargeted to the 22 joints of the
Sharpa Wave hand and sent to the hand driver, and the tracker pose is
converted by inverse kinematics on the YAM model into arm joint targets;
the operator watches the scene directly. A recorder running on the camera
clock ticks at 30\,Hz and, at every tick, stores the most recent commanded
arm and hand targets, the measured arm poses and hand joints, and the
latest tactile frame together with their wall-clock timestamps; the three
video streams are hardware-timestamped and aligned to the same ticks, so
that frame $k$ of every stream and row $k$ of the state log refer to the
same instant. Demonstrations are stored in LeRobot-v2 format, and ten
successful trajectories are kept per task.

\paragraph{Human demonstrations for fine-tuning}
The 90 human demonstrations per task in the fine-tuning mixture are not
taken from EgoForce but collected on the evaluation tasks themselves with a
lighter interface (\Cref{fig:hardware}, bottom right): the demonstrator
wears Wuji tactile gloves, a RealSense D435 camera on each wrist that
matches the robot's wrist views, and a VIVE Tracker 3.0 on each wrist. The
trackers give the wrist poses, the gloves give the finger joint angles and
fingertip tactile readings, and the wrist cameras give the two wrist
views, all time-stamped on the same clock as the teleoperation recorder.
The finger angles are retargeted to the 22 Sharpa Wave joints with the
same mapping as in teleoperation, the wrist poses are expressed as
delta-base end-effector poses, and the tactile readings are mapped to the
per-fingertip wrench inputs of the model, so that a human demonstration
enters fine-tuning in the same action and force space as a robot
demonstration.

\paragraph{Robot pretraining data}
The 100 hours of robot teleoperation data mixed into pretraining come from
T-Rex~\cite{trex}, collected on a Dexmate Vega-1 bimanual platform (7-DoF
arms) with the same Sharpa Wave hands and fingertip sensors, a ZED X Mini head
camera, and two ZED X One S wrist cameras at $640{\times}360$, teleoperated
with Manus gloves and VIVE trackers. Because arm actions are end-effector
poses rather than joint angles, the same 62-D action space covers both arm
embodiments.

\section{Implementation Details of Baselines}
\label{app:baselines}

\paragraph{$\pi_{0.5}$}
We use the official OpenPI implementation and the released $\pi_{0.5}$
base checkpoint, fine-tuned per task with full-parameter updates (no LoRA)
under FSDP on 8 GPUs. The policy takes the three camera views (head, left
wrist, right wrist), the task instruction, and the robot state; it has no
tactile input. Because the $\pi_{0.5}$ action head is fixed at 32
dimensions, the bimanual action is encoded as the 6-D end-effector pose of
each arm (position and roll--pitch--yaw, predicted as deltas) and a 10-D
PCA projection of each hand's 22 joints (absolute; the basis retains
96--97\% of the joint variance), and decoded back to 22 joints at
deployment. Action horizon 10. Training uses AdamW with gradient clipping
1.0, a cosine schedule from a peak learning rate of $2.5{\times}10^{-5}$
to $2.5{\times}10^{-6}$ over 20k steps with 200 warmup steps, global batch
size 32, and an EMA of the weights with decay 0.999.

\paragraph{T-Rex}
We use the official T-Rex implementation and start from its released
mid-trained checkpoint, which includes the tactile VQ-VAE and is the
recommended starting point for post-training. T-Rex shares the action
space of \model{}: the same 62-D delta-base end-effector pose plus 22
hand joints per arm and the same 16-step chunk, so our demonstrations are
used without any re-encoding. The model takes the three camera views at
$384{\times}288$, the instruction, and the 6-D fingertip wrenches of all
ten fingers through its VQ tactile encoder; the deformation-image branch is
disabled, since our hands provide no deformation images, and the FLARE
visual-prediction objective is kept on. Cascaded denoising uses 10 steps
split at step 6 with tactile dropout 0.1. Training uses AdamW with a peak
learning rate of $1{\times}10^{-4}$, cosine decay to zero without warmup,
weight decay 0, gradient clipping 1.0, per-device batch size 8 on 8 GPUs
with DeepSpeed ZeRO-2 in bf16, for 30 epochs.

\paragraph{\model{} ablations}
\emph{w/o force pred} removes the future-force tokens and their loss from the
\stagetwo{}, leaving the refiner to denoise the action suffix from the
observed tactile context only. \emph{w/o track pred} removes the track tokens,
the track heads, and the track losses from the \stageone{}, so coarse actions
are generated from video and state alone. Both variants are pretrained and
fine-tuned with the recipe of \Cref{tab:config}. The pretraining-recipe
variants of \Cref{tab:recipe} keep the architecture fixed: \emph{human w/o
force} trains on the same human--robot mixture with the human force
supervision disabled and the human tactile inputs masked,
and \emph{robot only} pretrains on the T-Rex robot data alone.

\section{Evaluation Tasks and Additional Results}
\label{app:results}

\begin{figure*}[!t]
\centering
\begin{tikzpicture}[lab/.style={anchor=south east, font=\scriptsize\bfseries, text=white, fill=black, fill opacity=0.5, text opacity=1, inner sep=2pt}]
\node[anchor=south west, inner sep=0] (h) at (0,0) {\includegraphics[height=7.2cm]{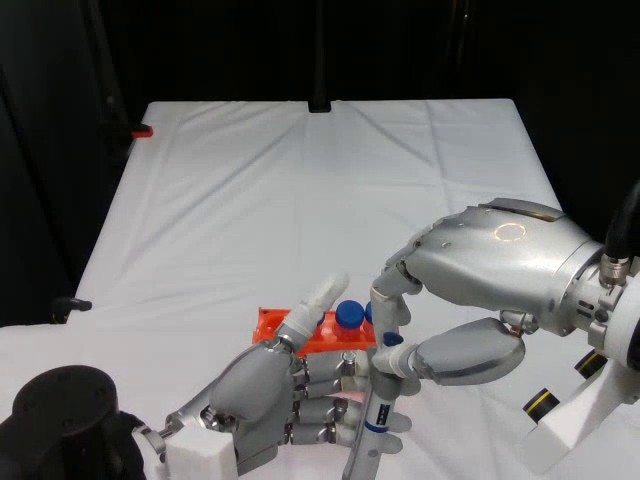}};
\node[anchor=north west, inner sep=0] (wl) at ([xshift=3pt]h.north east) {\includegraphics[width=4.55cm]{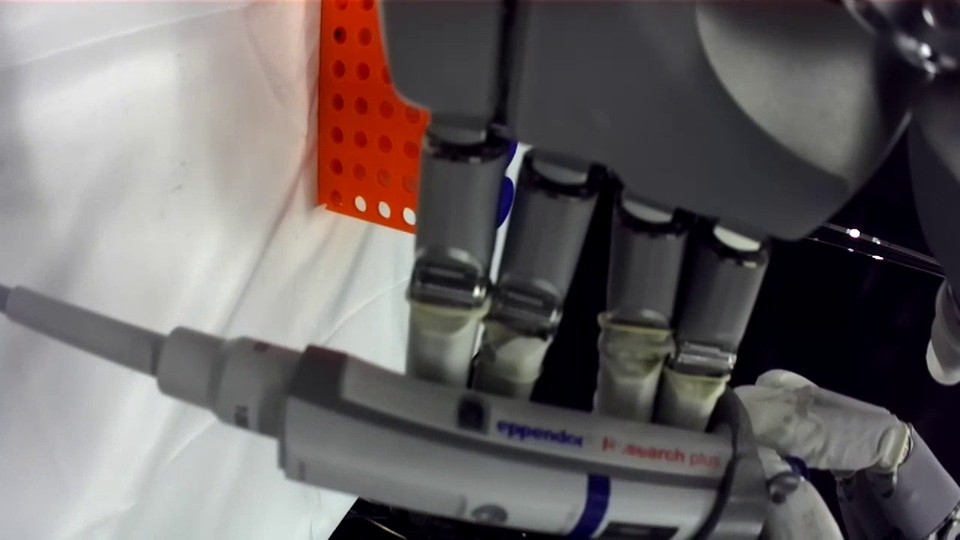}};
\node[anchor=north west, inner sep=0] (wr) at ([yshift=-3pt]wl.south west) {\includegraphics[width=4.55cm]{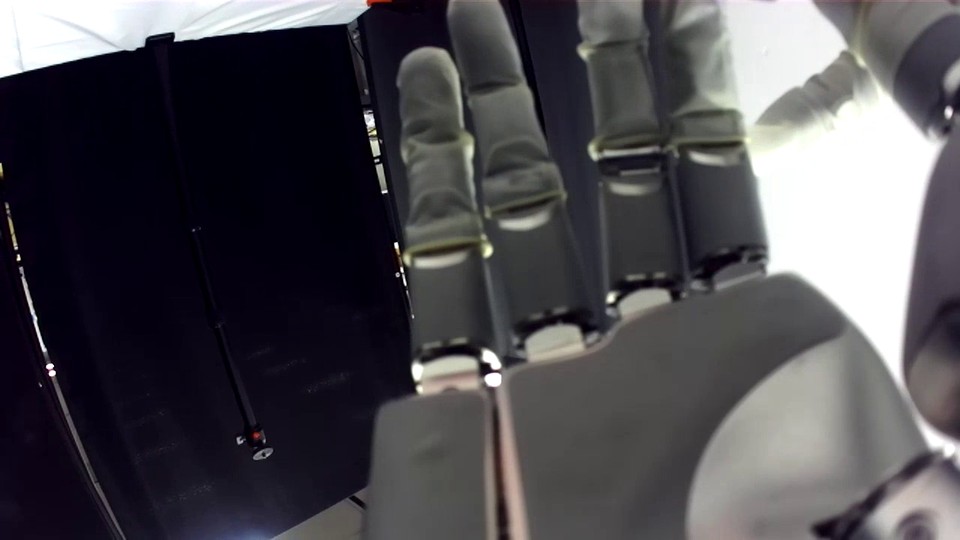}};
\node[anchor=north west, inner sep=0] (tc) at ([yshift=-3pt]wr.south west) {\includegraphics[width=4.55cm]{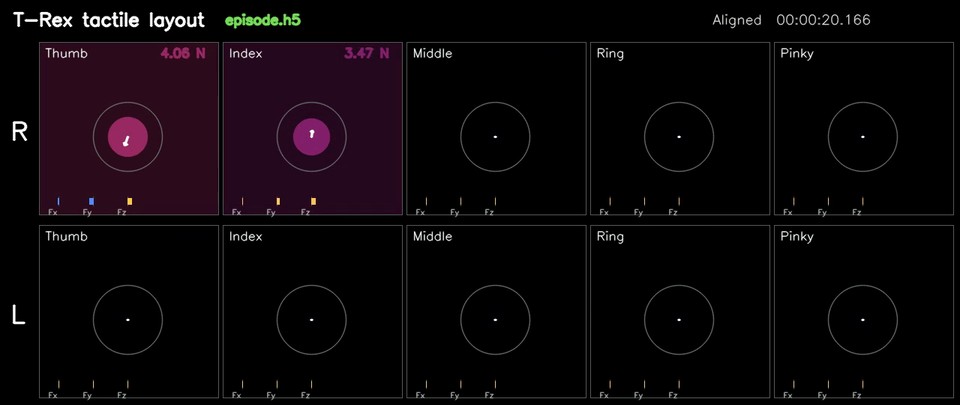}};
\node[lab] at (h.south east) {head view};
\node[lab] at (wl.south east) {left wrist view};
\node[lab] at (wr.south east) {right wrist view};
\node[lab] at (tc.south east) {tactile};
\end{tikzpicture}
\caption{\textbf{Full input modalities of \model{} on Task 10: Adjust
Pipette Volume.} \textit{Text Instruction:} ``A pipette set to
420\,$\mu$L is on the table. Hold the pipette body with your left hand and
turn the volume dial with your right thumb and index finger until the
display reads 400\,$\mu$L.'' Shown at one instant of an autonomous rollout
while the dial is being turned: the head-camera view, the left and right ZED
Mini wrist views, and the ten fingertip wrenches (rows R and L, one panel
per finger; the dot gives the tangential force and the color the normal
force), where the right thumb and index contacts on the dial appear while
the left hand holds the pipette body. Together with the robot state, these
are all the inputs the policy receives.}
\label{fig:inputs}
\vspace{36pt}   
\end{figure*}

\begin{figure*}[!tp]
\centering
\includegraphics[width=0.196\textwidth]{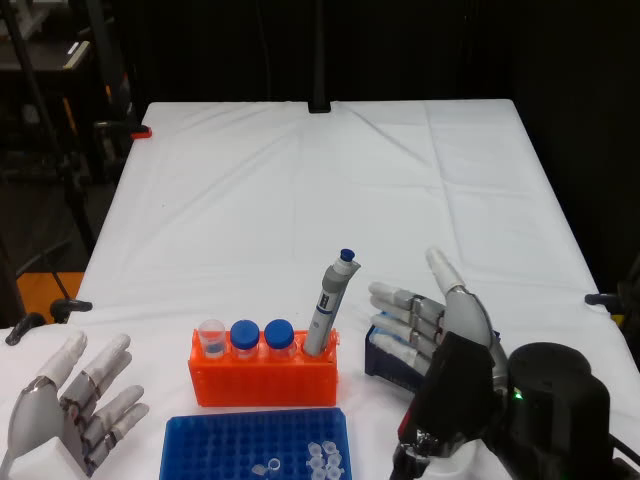}\hfill
\includegraphics[width=0.196\textwidth]{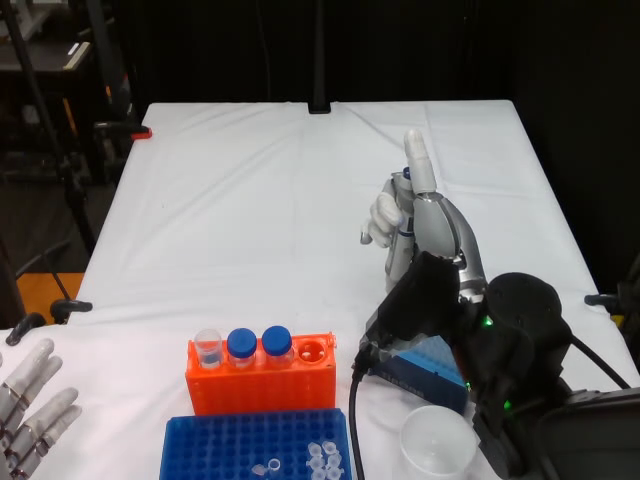}\hfill
\includegraphics[width=0.196\textwidth]{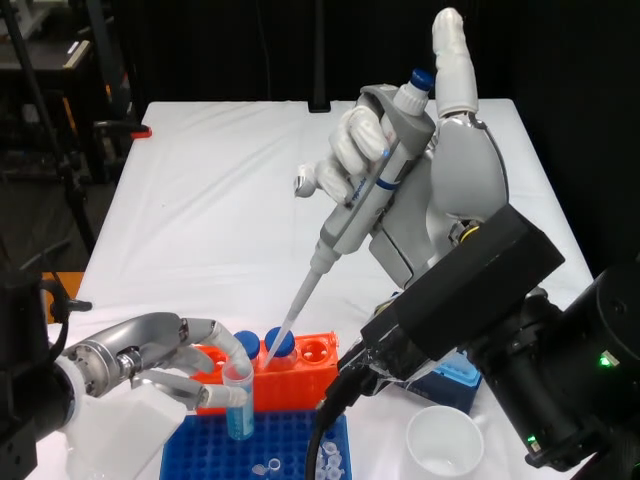}\hfill
\includegraphics[width=0.196\textwidth]{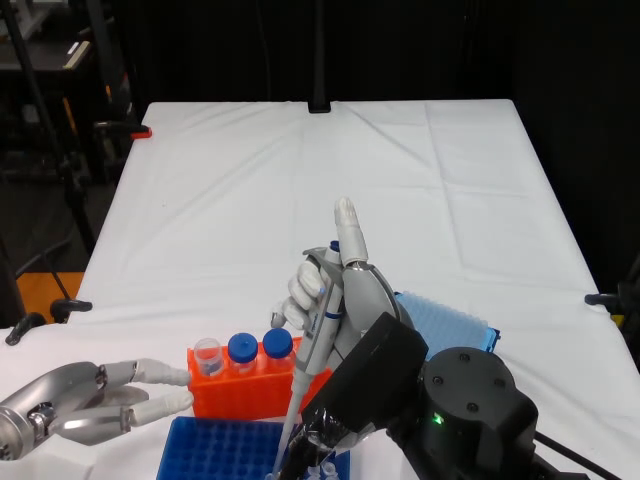}\hfill
\includegraphics[width=0.196\textwidth]{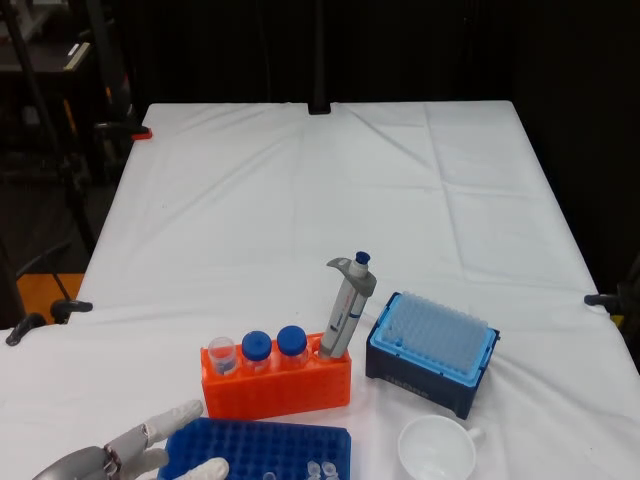}
\caption{Task 1: Lab Pipetting.}
\label{fig:task1}
\vspace{4pt}
\begin{minipage}{\textwidth}\footnotesize
\textbf{Task 1: Lab Pipetting.} \textit{Text Instruction:} ``An orange rack with a large tube and a small tube stands on the table, with a pipette, a box of tips, and a beaker beside it. Pick up the pipette with your right hand and press it into a tip. Lift the large tube from the rack with your left hand, draw liquid from it, and put the tube back. Then dispense the liquid into the small tube, eject the tip into the beaker, and put the pipette back down.'' The robot must press the pipette into a tip with the right hand, lift the large tube with the left hand and hold it steady, aspirate by pressing and releasing the plunger with the right index finger, return the tube to the rack, dispense into the small tube, eject the tip into the beaker, and put the pipette back down. A rollout succeeds only if liquid reaches the small tube, the tip is ejected, and the pipette is set back down, without dropping the pipette or the tube.
\end{minipage}
\end{figure*}

\begin{figure*}[!tp]
\centering
\includegraphics[width=0.196\textwidth]{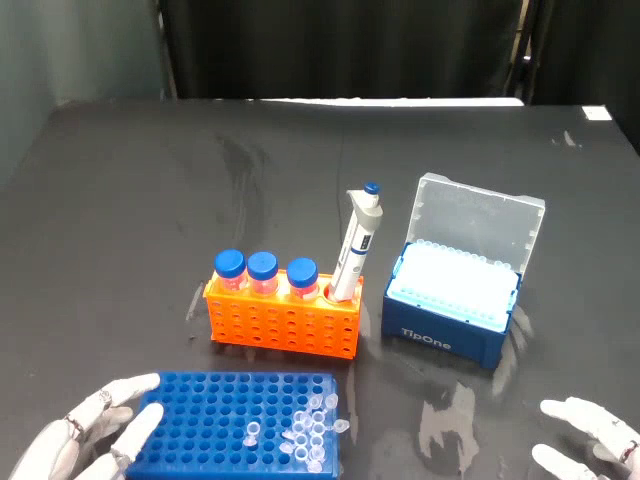}\hfill
\includegraphics[width=0.196\textwidth]{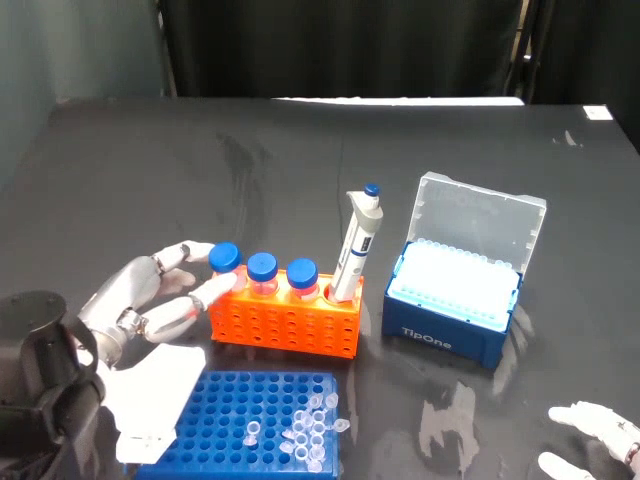}\hfill
\includegraphics[width=0.196\textwidth]{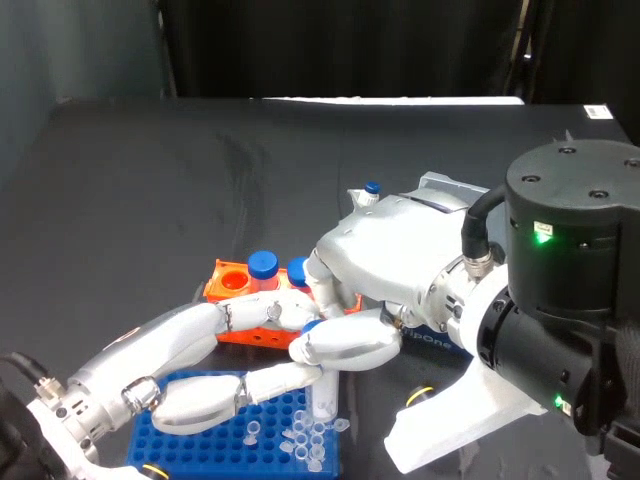}\hfill
\includegraphics[width=0.196\textwidth]{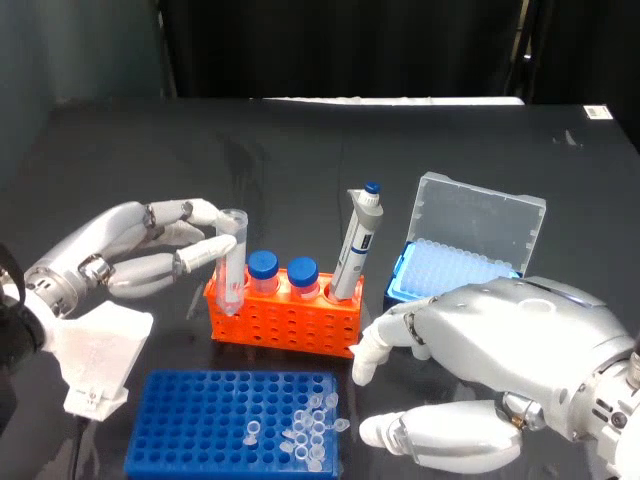}\hfill
\includegraphics[width=0.196\textwidth]{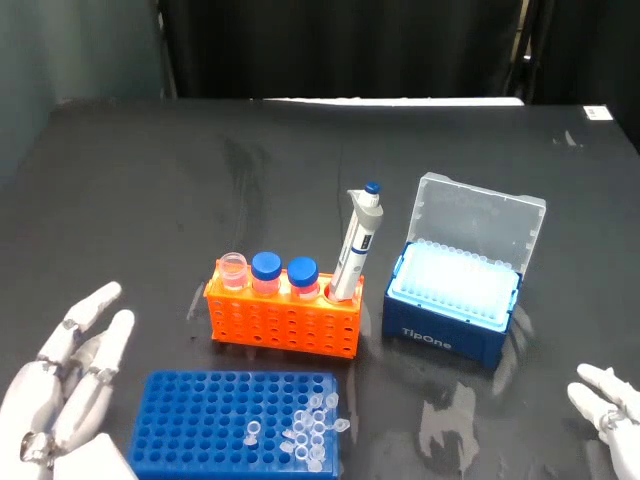}
\caption{Task 2: Open Test-Tube Cap.}
\label{fig:task2}
\vspace{4pt}
\begin{minipage}{\textwidth}\footnotesize
\textbf{Task 2: Open Test-Tube Cap.} \textit{Text Instruction:} ``An orange rack holding capped test tubes stands on the table in front of you. Pick up one tube with your left hand, unscrew its cap with your right hand, and place the tube back in the rack with your left hand.'' The robot must lift a capped tube from the rack with the left hand, unscrew the cap with the right thumb and index finger while the left hand holds the tube against the twist, and return the tube to the rack. A rollout succeeds only if the cap is fully separated and the tube stands in the rack at the end.
\end{minipage}
\end{figure*}

\begin{figure*}[!tp]
\centering
\includegraphics[width=0.196\textwidth]{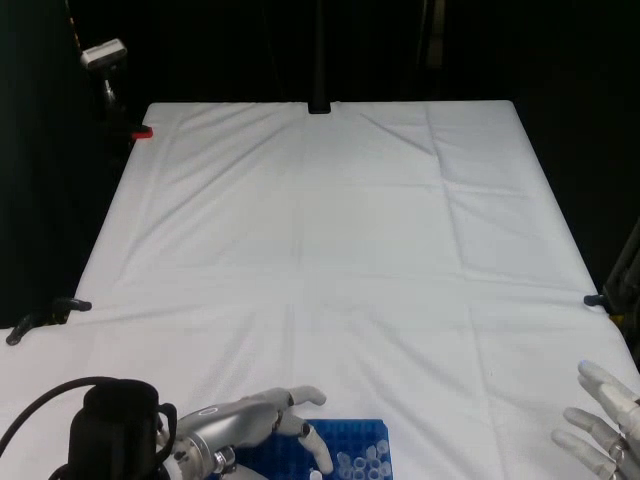}\hfill
\includegraphics[width=0.196\textwidth]{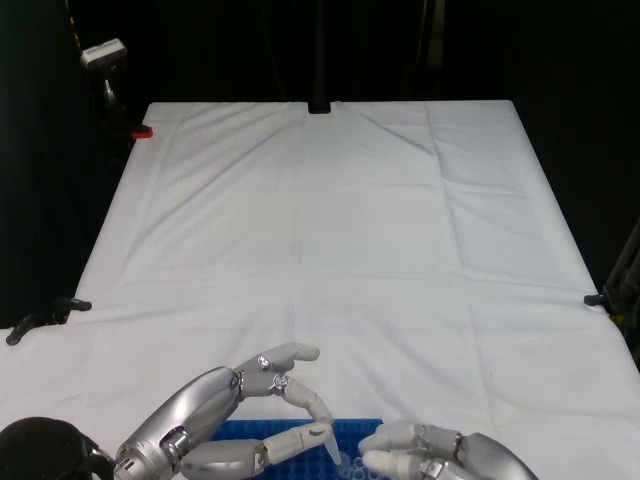}\hfill
\includegraphics[width=0.196\textwidth]{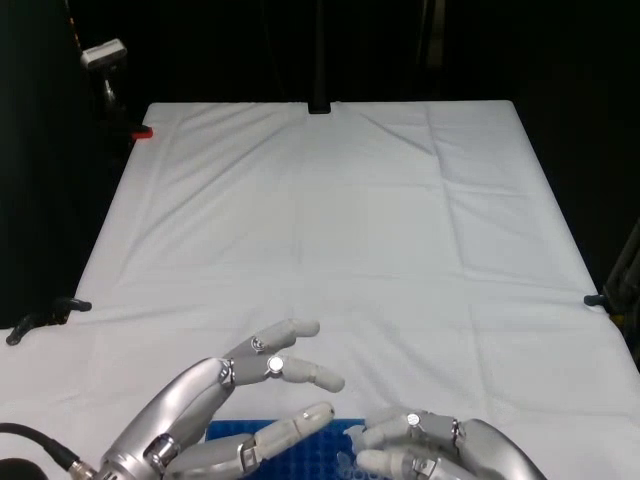}\hfill
\includegraphics[width=0.196\textwidth]{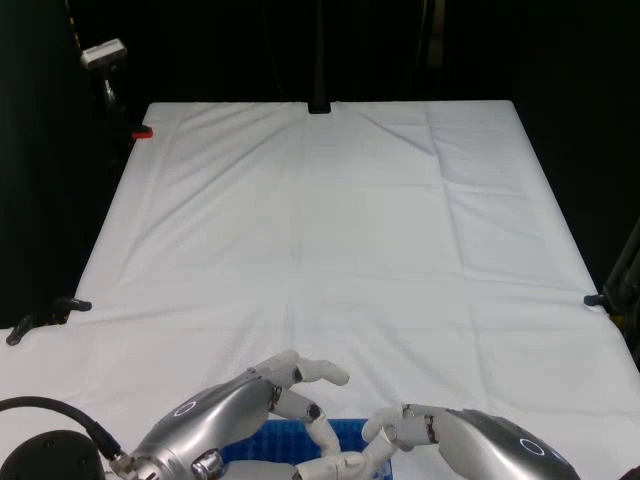}\hfill
\includegraphics[width=0.196\textwidth]{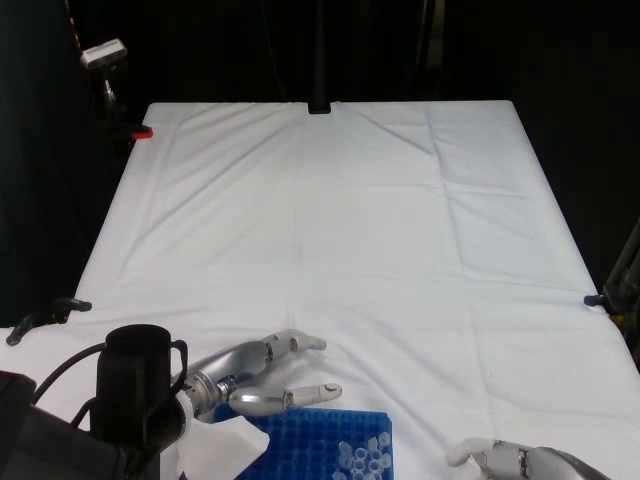}
\caption{Task 3: Open Microcentrifuge Flip Cap.}
\label{fig:task3}
\vspace{4pt}
\begin{minipage}{\textwidth}\footnotesize
\textbf{Task 3: Open Microcentrifuge Flip Cap.} \textit{Text Instruction:} ``A blue rack holding microcentrifuge tubes sits at the front edge of the table. Pick up one tube with your left hand, pass it to your right hand, and flip its hinged cap open with your left thumb; then return the tube to the rack.'' The robot must lift a tube from the rack with the left hand, pass it to the right hand, hook the left thumb under the lip of the cap and flip it open, and return the tube to the rack. A rollout succeeds only if the cap is fully open and the tube stands upright in the rack.
\end{minipage}
\end{figure*}

\begin{figure*}[!tp]
\centering
\includegraphics[width=0.196\textwidth]{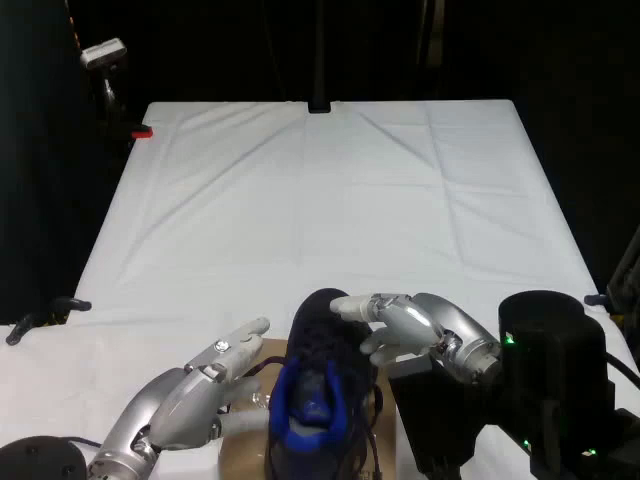}\hfill
\includegraphics[width=0.196\textwidth]{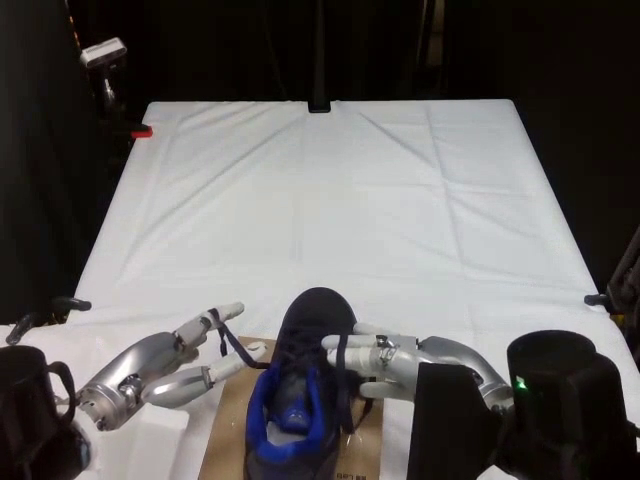}\hfill
\includegraphics[width=0.196\textwidth]{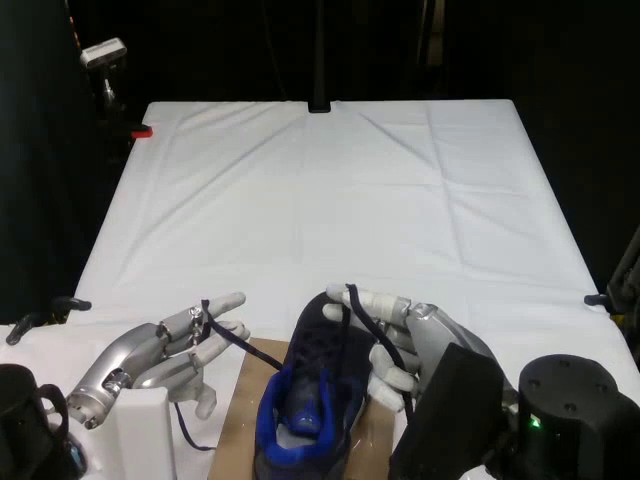}\hfill
\includegraphics[width=0.196\textwidth]{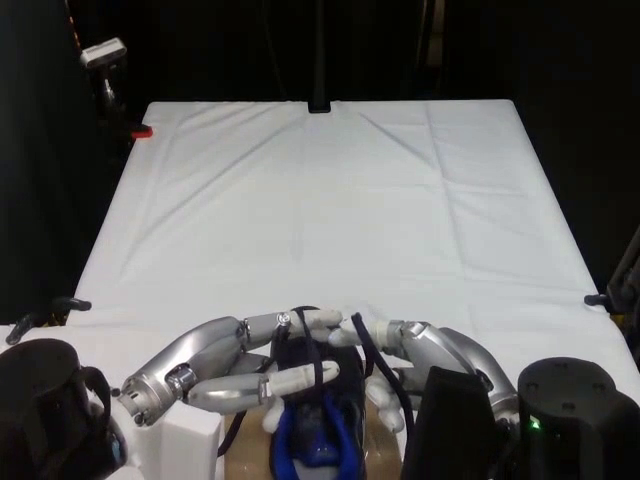}\hfill
\includegraphics[width=0.196\textwidth]{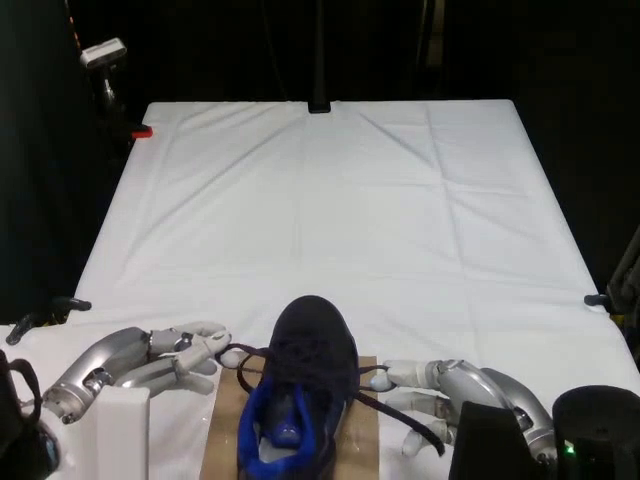}
\caption{Task 4: Tie Shoelaces.}
\label{fig:task4}
\vspace{4pt}
\begin{minipage}{\textwidth}\footnotesize
\textbf{Task 4: Tie Shoelaces.} \textit{Text Instruction:} ``A shoe with its laces untied sits on a board before you. Using both hands, pick up the two laces, cross and tighten them, form a loop with one lace, wrap the other lace around it, pull it through to make a second loop, and tighten the bow.'' The robot must cross and tighten the laces, form a loop, wrap the other lace around it, and pull both loops tight. A rollout succeeds only if a two-loop bow stays tied after the hands release.
\end{minipage}
\end{figure*}

\begin{figure*}[!tp]
\centering
\includegraphics[width=0.196\textwidth]{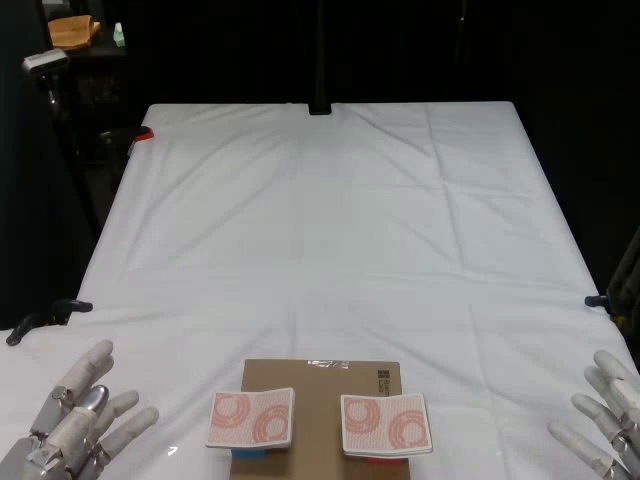}\hfill
\includegraphics[width=0.196\textwidth]{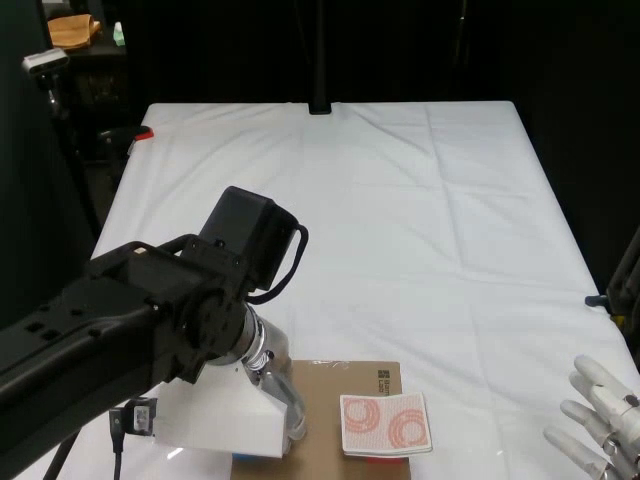}\hfill
\includegraphics[width=0.196\textwidth]{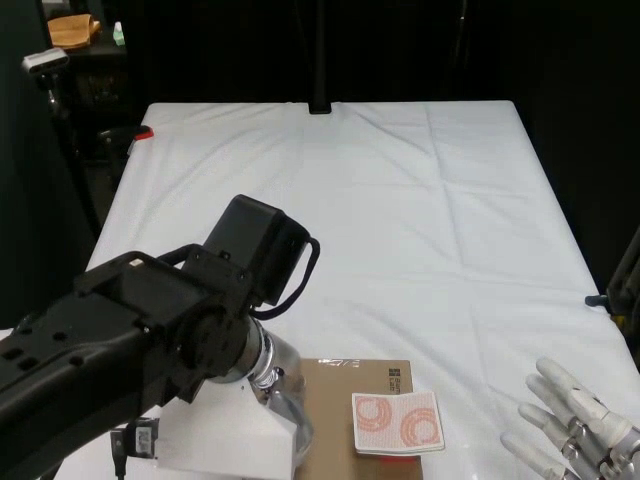}\hfill
\includegraphics[width=0.196\textwidth]{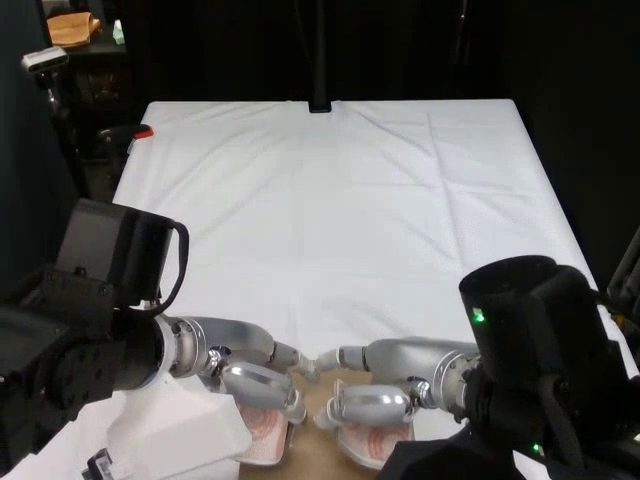}\hfill
\includegraphics[width=0.196\textwidth]{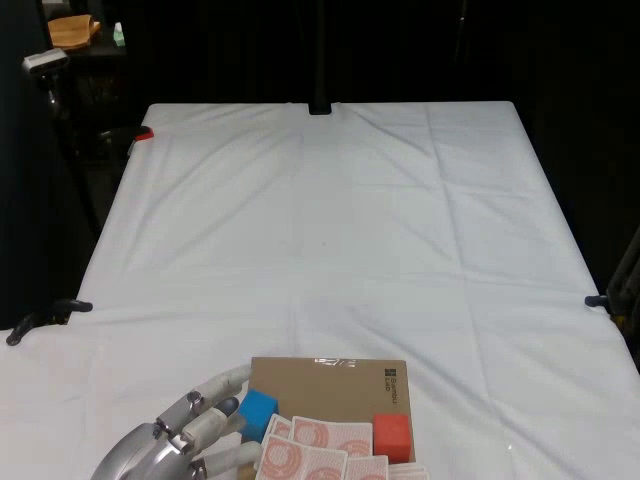}
\caption{Task 5: Shuffle Cards.}
\label{fig:task5}
\vspace{4pt}
\begin{minipage}{\textwidth}\footnotesize
\textbf{Task 5: Shuffle Cards.} \textit{Text Instruction:} ``Two halves of a deck of playing cards lie face down on a board in front of you. Pick up one half in each hand, bend the halves with your thumbs and riffle them together, then push the deck square.'' The robot must lift one half in each hand, bend both halves with the thumbs, release the cards so that they interleave, and push the deck square. A rollout succeeds only if the two halves merge into a single deck with no card dropped.
\end{minipage}
\end{figure*}

\begin{figure*}[!tp]
\centering
\includegraphics[width=0.196\textwidth]{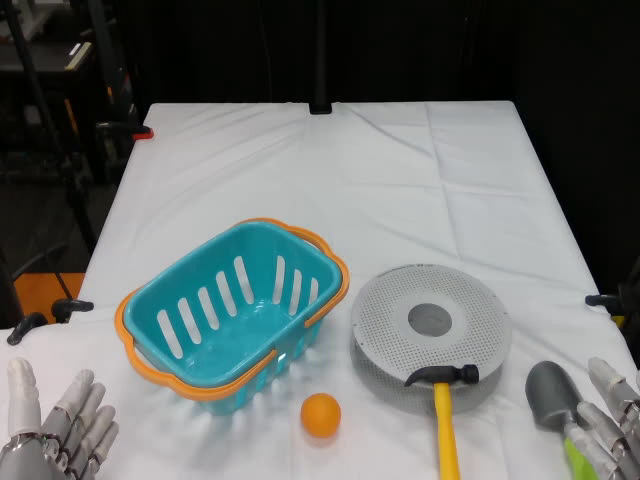}\hfill
\includegraphics[width=0.196\textwidth]{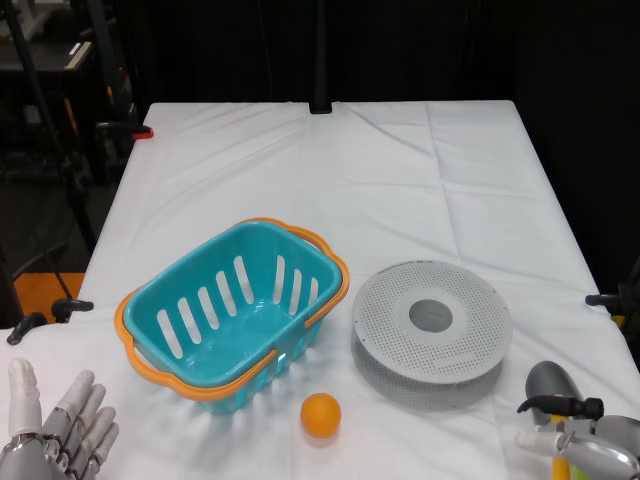}\hfill
\includegraphics[width=0.196\textwidth]{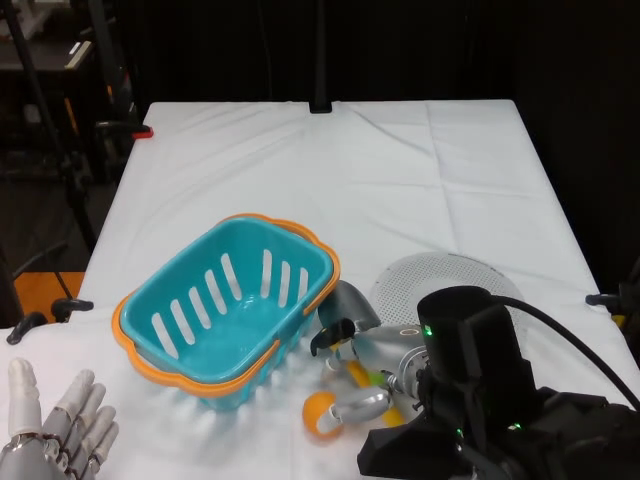}\hfill
\includegraphics[width=0.196\textwidth]{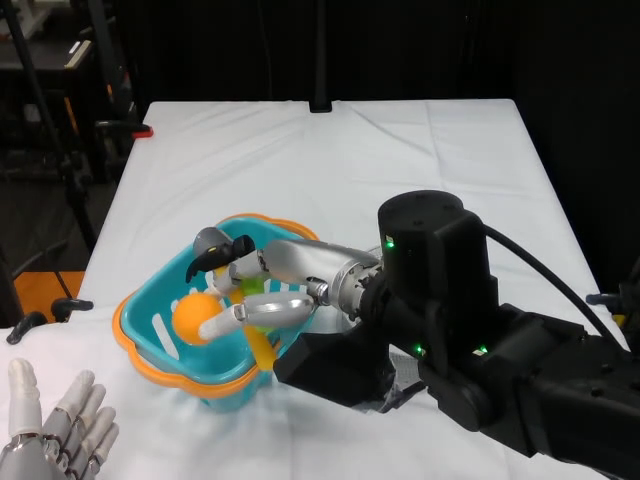}\hfill
\includegraphics[width=0.196\textwidth]{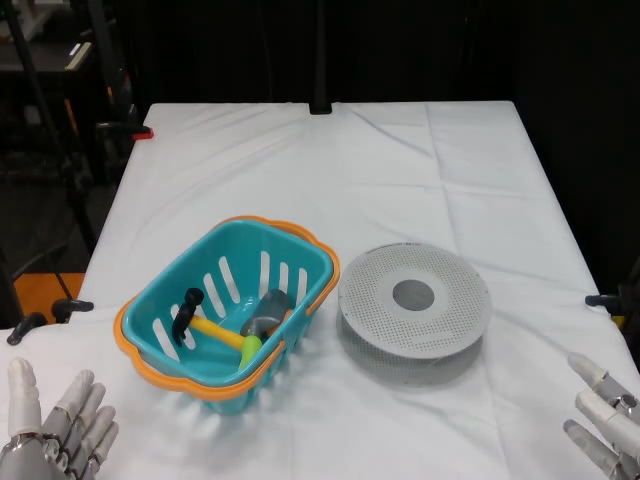}
\caption{Task 6: Multi-Object Grasp.}
\label{fig:task6}
\vspace{4pt}
\begin{minipage}{\textwidth}\footnotesize
\textbf{Task 6: Multi-Object Grasp.} \textit{Text Instruction:} ``A basket, a plate, a toy shovel, a toy hammer, and a ping-pong ball are on the table. With your right hand, pick up the toy shovel, then the toy hammer, then the ping-pong ball without releasing the objects already in hand, and place them into the basket.'' The robot must pick up the shovel, then the hammer and the ball with the same hand without releasing what it already holds, and place them into the basket. A rollout succeeds only if all three objects are held at the same time and end up in the basket.
\end{minipage}
\end{figure*}

\begin{figure*}[!tp]
\centering
\includegraphics[width=0.196\textwidth]{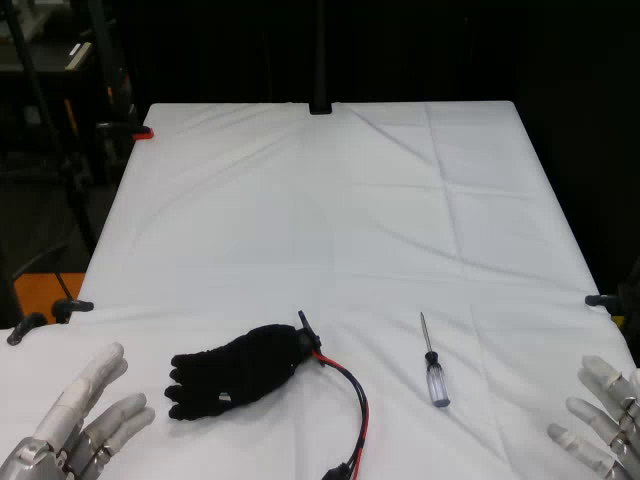}\hfill
\includegraphics[width=0.196\textwidth]{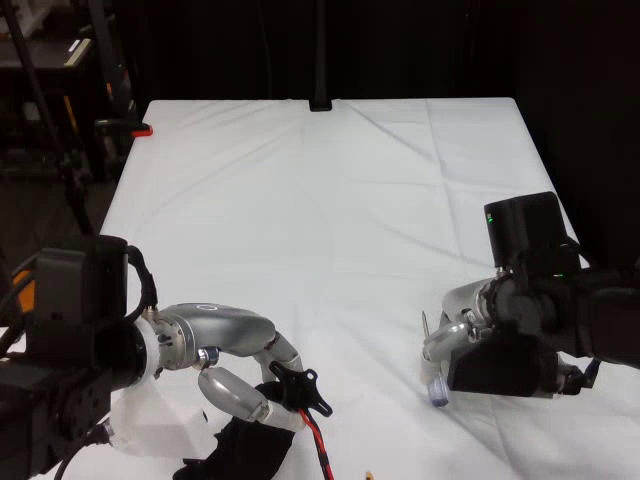}\hfill
\includegraphics[width=0.196\textwidth]{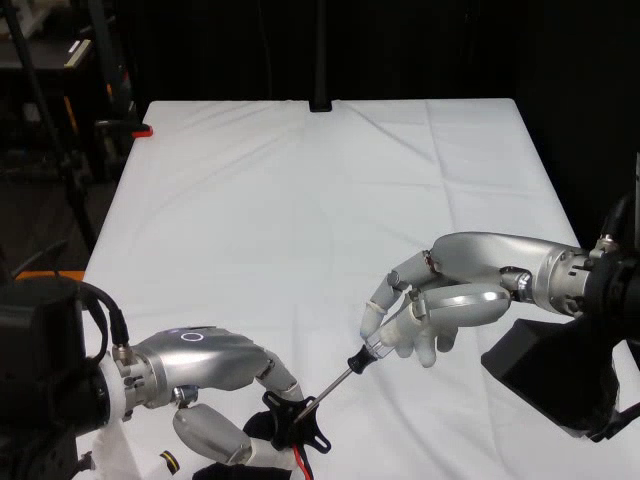}\hfill
\includegraphics[width=0.196\textwidth]{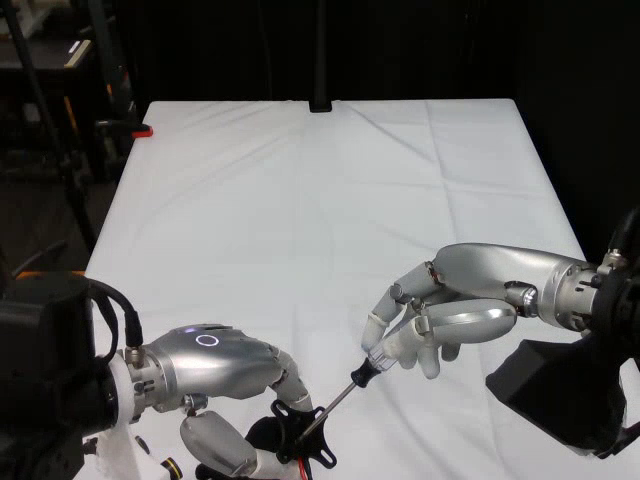}\hfill
\includegraphics[width=0.196\textwidth]{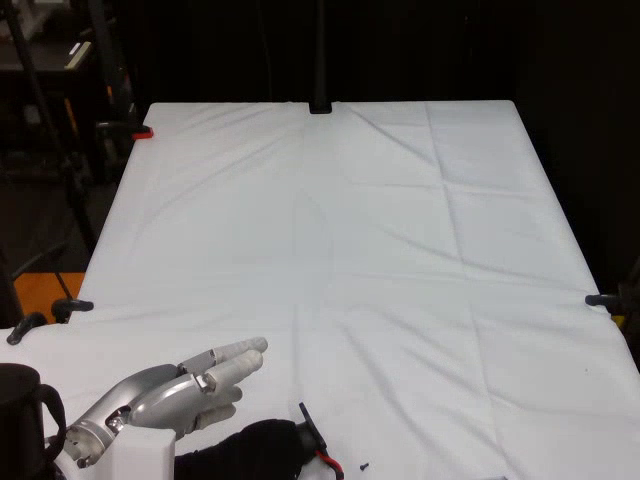}
\caption{Task 7: Use Screwdriver.}
\label{fig:task7}
\vspace{4pt}
\begin{minipage}{\textwidth}\footnotesize
\textbf{Task 7: Use Screwdriver.} \textit{Text Instruction:} ``A Wuji Hand lies on the left of the table and a screwdriver on the right. Hold the Wuji Hand up with your left hand, base facing you, pick up the screwdriver with your right hand, seat its tip in the central screw, and turn it counterclockwise until the screw is out; then put both back on the table.'' The robot must hold the Wuji Hand steady, seat the screwdriver tip in the screw with light axial pressure, and turn it counterclockwise with repeated regrasps while the left hand resists the torque. A rollout succeeds only if the screw is removed and both objects are back on the table.
\end{minipage}
\end{figure*}

\begin{figure*}[!tp]
\centering
\includegraphics[width=0.196\textwidth]{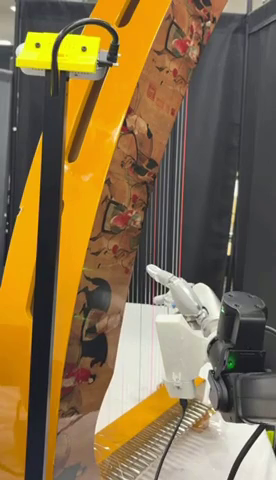}\hfill
\includegraphics[width=0.196\textwidth]{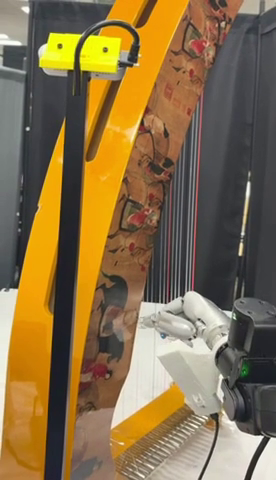}\hfill
\includegraphics[width=0.196\textwidth]{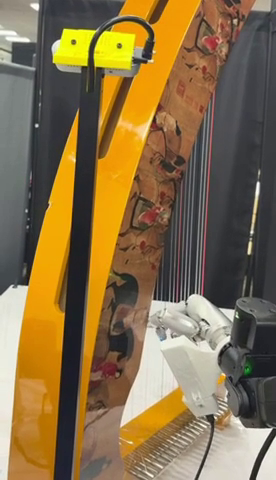}\hfill
\includegraphics[width=0.196\textwidth]{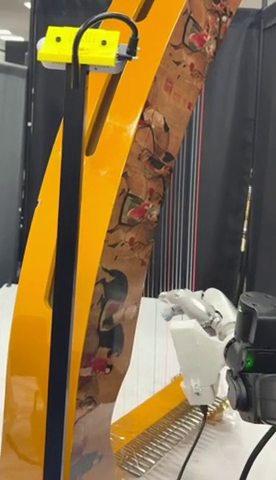}\hfill
\includegraphics[width=0.196\textwidth]{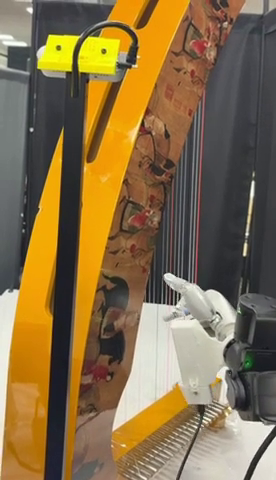}
\caption{Task 8: Play Harp.}
\label{fig:task8}
\vspace{4pt}
\begin{minipage}{\textwidth}\footnotesize
\textbf{Task 8: Play Harp.} \textit{Text Instruction:} ``A harp stands in front of you. With your right hand, hook a fingertip behind each of the ten indicated strings in turn and pluck it so that it sounds.'' The robot must bring the right hand to the strings and, for each of the ten indicated strings, hook a fingertip behind it and release it with enough force to make the string ring cleanly, neither muting it nor snapping it. The score is the number of the ten notes that sound cleanly.
\end{minipage}
\end{figure*}

\begin{figure*}[!tp]
\centering
\includegraphics[width=0.196\textwidth]{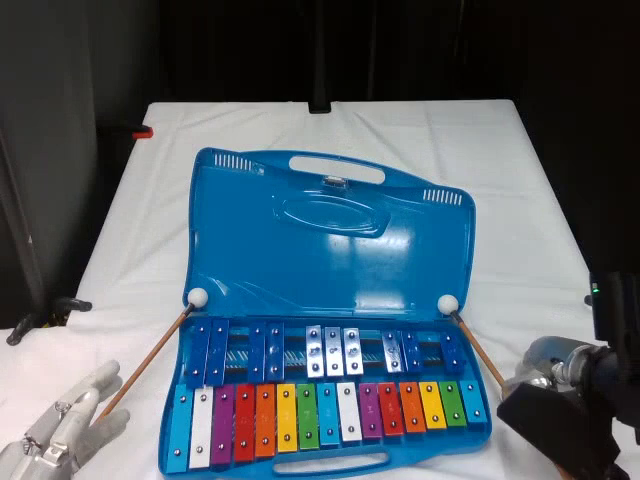}\hfill
\includegraphics[width=0.196\textwidth]{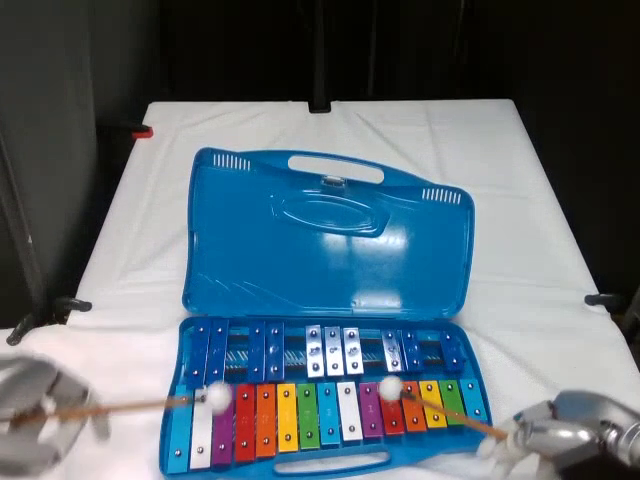}\hfill
\includegraphics[width=0.196\textwidth]{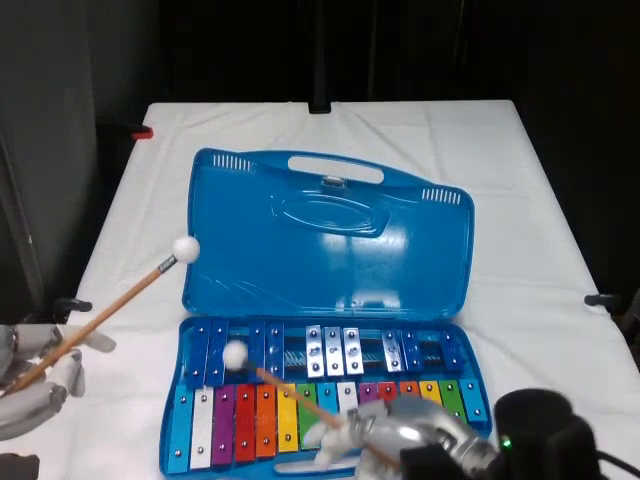}\hfill
\includegraphics[width=0.196\textwidth]{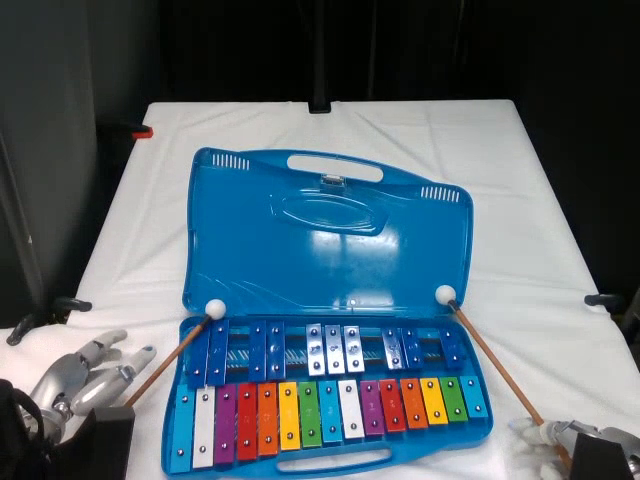}\hfill
\includegraphics[width=0.196\textwidth]{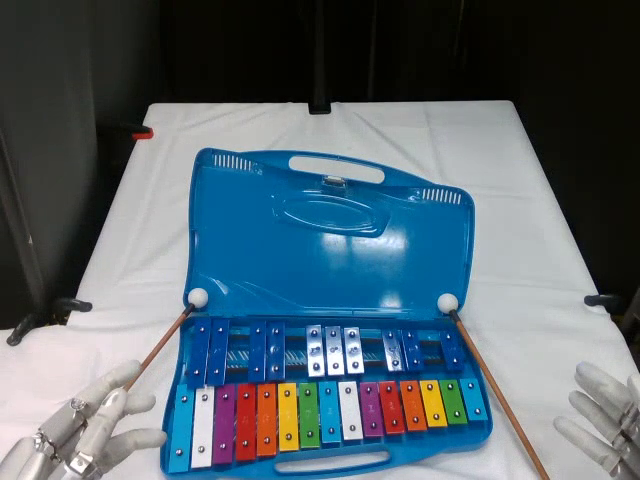}
\caption{Task 9: Play Xylophone.}
\label{fig:task9}
\vspace{4pt}
\begin{minipage}{\textwidth}\footnotesize
\textbf{Task 9: Play Xylophone.} \textit{Text Instruction:} ``A xylophone lies open on the table with a mallet beside it. Pick up the mallet with your right hand and strike each of the ten indicated bars in turn so that it rings.'' The robot must pick up the mallet with the right hand, hold it in a stable grip, and strike each of the ten indicated bars with enough force to ring cleanly, neither a muted tap nor a bounced double hit. The score is the number of the ten notes that sound cleanly.
\end{minipage}
\end{figure*}

\begin{figure*}[!tp]
\centering
\includegraphics[width=0.196\textwidth]{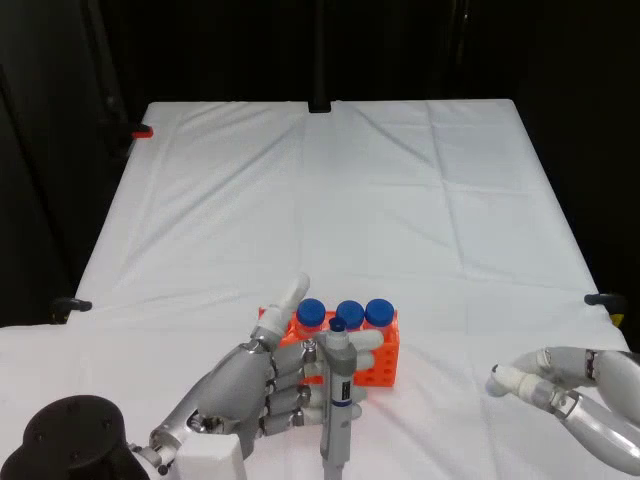}\hfill
\includegraphics[width=0.196\textwidth]{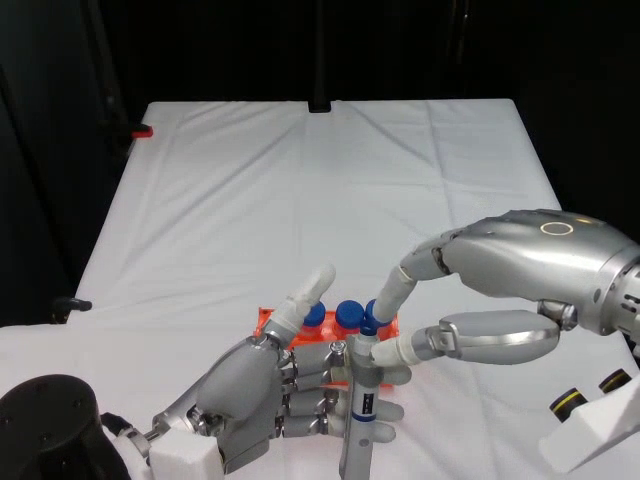}\hfill
\includegraphics[width=0.196\textwidth]{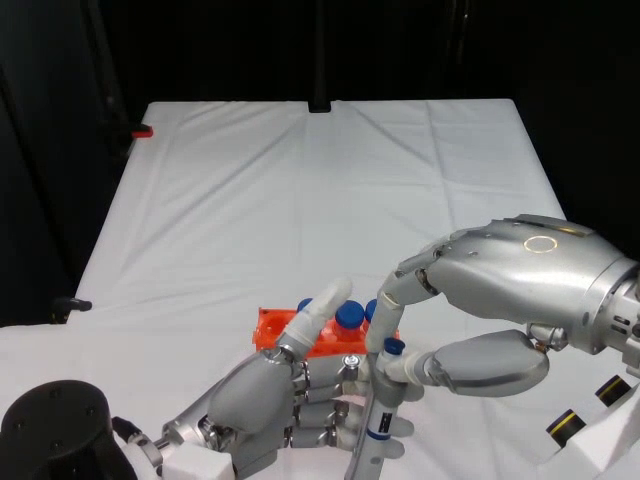}\hfill
\includegraphics[width=0.196\textwidth]{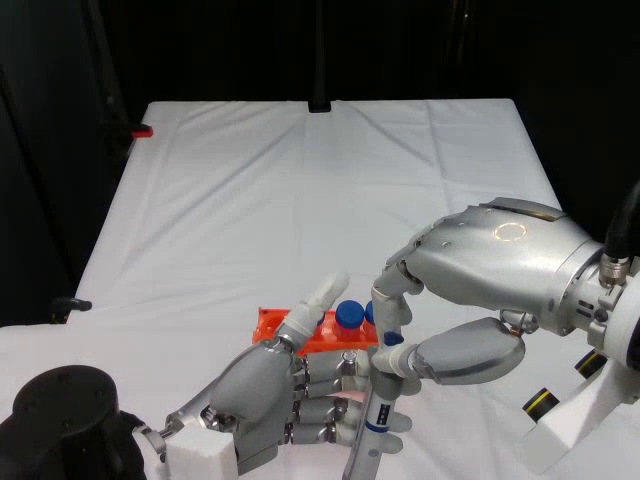}\hfill
\includegraphics[width=0.196\textwidth]{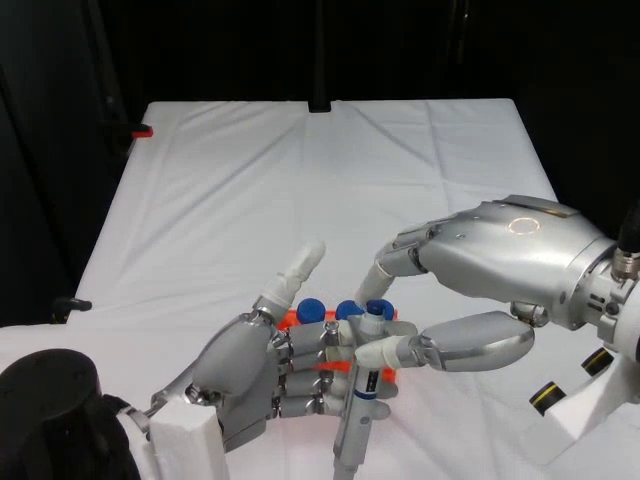}
\caption{Task 10: Adjust Pipette Volume.}
\label{fig:task10}
\vspace{4pt}
\begin{minipage}{\textwidth}\footnotesize
\textbf{Task 10: Adjust Pipette Volume.} \textit{Text Instruction:} ``A pipette set to 420\,$\mu$L is on the table. Hold the pipette body with your left hand and turn the volume dial with your right thumb and index finger until the display reads 400\,$\mu$L.'' The robot must hold the pipette body with the left hand, roll the volume dial with the right thumb and index fingertip in small increments from 420 toward 400\,$\mu$L, and stop when the display reads 400. A trial counts as engaged if the dial is turned; we report the engaged trials of ten and, over them, the mean absolute error of the final reading from 400\,$\mu$L.
\end{minipage}
\end{figure*}

We evaluate on ten contact-rich tasks in three classes. \emph{Binary tasks}
(Tasks 1--7) require a sequence of grasps and in-hand adjustments and are
scored 1 on success and 0 otherwise. \emph{Continuous tasks} (Tasks 8 and 9)
require repeated contacts with the right amount of force: the policy
plucks each of ten indicated harp strings or strikes each of ten indicated
xylophone bars, one note per position, and each rollout is scored by the
number of these ten notes that sound cleanly, so partial progress is
credited. A note counts only if it sounds cleanly: a pluck or strike that
is too light produces no sound, and one that is too hard snaps or buzzes
the string or bounces the mallet into a double hit. The target is not a
melody but a clean sound at each indicated position. The \emph{volume-adjustment} task (Task 10) requires fine dial
manipulation; we report the number of trials (of ten) in which the dial is
engaged and, over those trials, the mean absolute error of the final reading
from the 400\,$\mu$L target, since a never-engaged trial is a failure with
no reading. Every task is evaluated over ten rollouts with object placement
randomized per rollout (\Cref{fig:randomness}), and all methods share the
same scenes and objects. Each task figure below is followed by the task's text instruction, of which
the prompt used in fine-tuning is a one-line abbreviation (e.g., ``Play
harp.''), and by the behavior required for success;
\Cref{fig:inputs} shows the full input modalities of the policy on one task:
the text instruction, the three camera views, and the fingertip wrenches at
one instant of a rollout.
\Cref{fig:task1,fig:task2,fig:task3,fig:task4,fig:task5,fig:task6,fig:task7,fig:task8,fig:task9,fig:task10}
show key stages of one rollout per task (five frames sampled uniformly
from the head camera, an external camera for the harp; for Task 7 the
frames come from a teleoperated demonstration).

\paragraph{Failure modes}
The failures we observe in the evaluation rollouts fall into three groups,
all rooted in contact rather than in gross motion.
\emph{Lost contact without feedback.} In shoelace tying the critical moment
is the first pinch: if the fingertips close beside the lace instead of on
it, nothing in the images reveals the miss, and a policy without force
feedback continues the crossing and looping motion open-loop on an empty
hand, so every later step fails. Contact information is what allows a
re-grasp here; the variant trained without force prediction never ties a
bow (0\%), whereas the full model recovers in some rollouts.
\emph{Force precision.} Pipetting, cap opening, and card shuffling fail at
the level of how hard the fingers press rather than where they go. Pressing
the plunger too hard past the first stop aspirates the wrong volume or air,
while pressing too lightly leaves the liquid in the tip; the cap tasks fail
when the pinch does not transmit enough torque to the cap; and in card
shuffling too much thumb force bends the halves until cards fly out, while
too little lets the half-deck slip out of the grip before the riffle. The
harp is the extreme case: a pluck that is too forceful snaps or detunes a
string, and one that is too light produces no note, so the score depends
almost entirely on regulating fingertip force on a compliant object.
\emph{Dexterity under contact.} In screwdriver use the tip cams out of the
screw head when the axial force drops during a turn, and the Wuji Hand slips in
the left hand when its grip does not resist the torque; in multi-object
grasping the first object is dropped while the fingers reconfigure for the
second, and in pipette-volume
adjustment the dial is never engaged in half of the trials; once engaged,
four of five readings end within 4\,$\mu$L of the target, so the failure is
in committing to the contact rather than in regulating it.
These groups separate the ablations: removing force prediction produces the
open-loop and force-precision failures, while removing track prediction
mostly produces misplaced but force-appropriate motions.

\setlength{\abovecaptionskip}{2pt}

\end{document}